\documentclass[preprint,12pt]{elsarticle}

\usepackage{amssymb}
\usepackage{amsmath}
\usepackage{booktabs}
\usepackage{multirow}
\usepackage{xcolor}
\usepackage{url}
\usepackage{lineno}

\journal{Engineering Applications of Artificial Intelligence}

\begin{document}

\begin{frontmatter}

\title{A Channel-Boosted Multi-Agent System with Iterative Consultation for Document Sensitivity Classification}

\author[pr,dl]{Aleesha Zainab}

\author[pr,paic,dl]{Asifullah Khan\corref{cor1}}
\ead{asif@pieas.edu.pk}
\cortext[cor1]{Corresponding author.}

\author[pr]{Muhammad Ahmed Khalid}

\author[pr]{Faheem Ullah Khan}

\affiliation[pr]{organization={Pattern Recognition Lab, Department of Computer and Information Sciences (DCIS), Pakistan Institute of Engineering and Applied Sciences (PIEAS)},
            city={Nilore, Islamabad},
            country={Pakistan}}

\affiliation[paic]{organization={PIEAS Artificial Intelligence Center (PAIC), Pakistan Institute of Engineering and Applied Sciences (PIEAS)},
            city={Nilore, Islamabad},
            country={Pakistan}}

\affiliation[dl]{organization={Deep Learning Lab, Center for Mathematical Sciences, Pakistan Institute of Engineering and Applied Sciences (PIEAS)},
            city={Nilore, Islamabad},
            country={Pakistan}}

\begin{abstract}
Organizations responsible for critical national infrastructure must identify sensitive
documents before they are stored, shared, or transmitted. Manual review is time-consuming
and can be inconsistent, particularly as document volumes increase. Transformer-based
classifiers offer an alternative, but fixed input limits can exclude important evidence
from long documents, which are often sensitive. The artificial intelligence contribution
is the Channel-Boosted Multi-Agent System (CB-MAS), a paradigm in which specialized agents
generate, dynamically weight, and fuse complementary representation channels, instantiated as
the Iterative Consultation Multi-Agent System (IC-MAS). In IC-MAS, a Channel Critic Agent learns
document-specific trust weights that gate the fusion of BERT and a Robustly Optimized BERT
Approach (RoBERTa) encoder on a document's first window, while two Consultation Agents,
initialized from the first and last windows, iteratively exchange belief states and stop once
confident. A Decision and Verification Agent uses a small language model to resolve
disagreements with a full-document Domain Evidence Agent. On Strategic~16K, a leakage-controlled
corpus of 16{,}000 publicly released diplomatic cables, 5-fold cross-validation gives the core
system 90.72\% accuracy, an F1-score of 0.9123, and sensitive-class recall of 0.9201, against 89.14\%,
89.33\%, and 86.66\% for the strongest single-encoder baseline (McNemar's test,
$p=5.15\times10^{-14}$), while using 54\% fewer consultation rounds than a fixed-round variant.
The full system raises sensitive recall to 93.54\%. Ablations attribute most of the gain to
critic-controlled fusion, and post-hoc explanations support auditable deployment.
\end{abstract}

%% Graphical abstract: uncomment and supply an image (e.g. the system schematic) if the
%% EAAI submission system asks for one.
%\begin{graphicalabstract}
%\includegraphics[width=\linewidth]{fullpic.png}
%\end{graphicalabstract}

\begin{highlights}
\item Channel-Boosted MAS fuses BERT and RoBERTa via adaptive critic weights.
\item Two agents read a document's first and last windows and iteratively consult.
\item Adaptive halting cuts consultation rounds by 54\% with no loss in accuracy.
\item Sensitive recall rises from 86.66\% (BERT) to 93.54\% with the full system.
\item Ablations, negative results, McNemar tests and LIME/SHAP explanations reported.
\end{highlights}


\end{frontmatter}

%% Uncomment for line numbers (often requested at the review stage):
%\linenumbers

\section{Introduction}\label{sec:intro}

Organizations in all sectors are creating and accumulating more digital documents than ever
before, and a significant portion of that content is sensitive: information that, if disclosed,
could lead to financial, operational, or reputational damage. Identifying, categorizing, and
safeguarding such information before it is accidentally or maliciously disclosed is now a
pressing operational challenge.

Misclassification is particularly costly for organizations in critical national infrastructure
sectors. In many such institutions, trained staff still manually review documents to determine
their sensitivity level before they are routed, stored, or transmitted, a process that is slow,
inconsistent, and hard to scale as document volumes grow \cite{zainab2026benchmark}. More
broadly, the IBM \textit{Cost of a Data Breach Report 2023} puts the global average cost of a
single data breach at \$4.45~million, an increase of 15.3\% from 2020 \cite{ibm2023}.
% TODO(author): consider citing the most recent IBM report (2025) instead.
In sectors such as energy and defense, the consequences extend beyond financial loss to
national security, public safety, and institutional reputation.

Commercial Data Loss Prevention (DLP) tools typically rely on rule-based mechanisms such as
keyword lists and hash matching \cite{hart2011}, which are too coarse to capture the contextual
features that make a document sensitive. Existing AI-based solutions, in turn, largely treat
classification as a black box and offer no explanation for why a label was assigned
\cite{rudin2019}, a serious shortcoming in regulatory and security contexts.

Our earlier work \cite{zainab2026benchmark} addressed the first of these limitations by
benchmarking classical machine learning and transformer-based models (BERT, RoBERTa, ELECTRA)
on a purpose-built, leakage-controlled corpus. The present paper addresses a structural
limitation surfaced by that benchmark, truncation-induced evidence loss in long documents. We
design, implement, and validate a modular multi-agent system (MAS), in which multiple
specialized, interacting computational agents jointly solve a task that no single agent solves
alone, together with a post-hoc explainable AI (XAI) mechanism for interpretable sensitivity
classification. The agents in the core classifier are trainable encoders and lightweight
neural modules rather than prompted large language models; Section~\ref{subsec:overview}
defines the sense of ``agent'' used throughout.

\subsection{Problem Statement}\label{sec:problem}

Automated, scalable, and explainable document sensitivity classification remains an open need
for organizations that handle sensitive documents. Five technical gaps motivate this work:
(1)~\textit{semantic understanding}: rule-based and classical ML systems miss contextual
meaning; (2)~\textit{explainability}: existing systems lack transparency;
(3)~\textit{multi-agent architecture}: most existing systems consist of a single monolithic
model, and there is no principled way to enrich a classifier's context by generating,
dynamically weighting, and fusing multiple complementary agent channels;
(4)~\textit{scalability and generalisability}: existing frameworks are domain-specific; and
(5)~\textit{truncation-induced evidence loss}: transformer baselines process only a
fixed-length input window and discard evidence outside it, precisely in the longer, sensitive
documents for which correct detection matters most. Our prior work \cite{zainab2026benchmark}
addressed gap~(1) through a leakage-controlled transformer benchmark. This paper addresses
gaps~(3) and~(5) directly and gap~(2) through a post-hoc explainability mechanism; gap~(4) is
discussed in Section~\ref{sec:limitations}.

For gap~(3) we propose Channel-Boosted MAS (CB-MAS), a multi-agent paradigm in which a
system's context is enriched by generating, dynamically weighting, and fusing multi-source
agent channels, rather than by a single monolithic model or a fixed, hand-crafted combination
rule. IC-MAS, the architecture developed and evaluated in this paper, is a concrete
instantiation of CB-MAS that targets gap~(5): its Channel Critic Agent dynamically weights
channels from two complementary encoders of a document's first window, which are then fused
into a single representation, while its Consultation Agents iteratively update belief states
initialized from the beginning and end of the document, reconciling the two windows through
message passing rather than a one-shot fusion step.

A more specific issue underlies gap~(3). In principle, combining two pretrained encoders
trained under different objectives (e.g., BERT and RoBERTa) should help, since each may
capture signal the other misses. Without a mechanism for deciding, per document, how much to
trust each encoder, however, a fixed combination rule (averaging, majority voting, or always
trusting one encoder) can do more harm than good. Section~\ref{sec:fusionablation} confirms
this empirically: raw max-pool fusion of BERT and RoBERTa, with no learned weighting, reaches
only 88.74\% accuracy, \emph{below} the BERT-only baseline of 89.14\%
(Section~\ref{subsec:baselineresults}). Channel Boosting addresses this by having a Channel
Critic Agent learn, per document, a pair of trust weights that determine how strongly each
encoder's representation is amplified by the other, turning the combination into a learned
function rather than a fixed rule; Section~\ref{subsec:corecomparison} shows that this
mechanism accounts for most of the improvement over the single-encoder baseline. The separate,
window-level problem of reconciling evidence from the beginning and end of a document is
handled by the Consultation Agents (Section~\ref{sec:iterconsult}) and examined in
Section~\ref{sec:whyconsult}.

\subsection{Contributions}\label{sec:contrib}

This work makes the following contributions:
\begin{itemize}
\item We introduce Channel-Boosted MAS (CB-MAS) and develop IC-MAS for document sensitivity classification. A Channel Critic learns how to combine BERT and RoBERTa representations, while two agents exchange information from the beginning and end of each document.
\item We add a Decision \& Verification Agent to examine disagreements between the window-based classifier and a full-document evidence model. This step improves sensitive-document detection on long documents.
\item We evaluate the architecture through controlled ablations, including unsuccessful variants, to identify which components contribute to performance.
\item We compare IC-MAS with the single-encoder benchmark of \cite{zainab2026benchmark}, assess statistical significance, and provide post-hoc explanations for reviewing predictions.
\end{itemize}

The remainder of this paper is organized as follows. Section~\ref{sec:relwork} reviews related
work. Section~\ref{sec:dataset} summarizes the dataset and leakage-removal protocol underlying
the Strategic~16K corpus, first introduced in \cite{zainab2026benchmark}.
Section~\ref{sec:baseline} recaps the baseline results against which the proposed architecture
is compared. Section~\ref{sec:architecture} presents the proposed architecture, and
Section~\ref{sec:methodology} describes the experimental methodology.
Section~\ref{sec:results} presents results and ablation studies. Section~\ref{sec:stats}
reports statistical validation and explainability. Section~\ref{sec:discussion} discusses the
design, its multi-agent properties, limitations, and future work, and
Section~\ref{sec:conclusion} concludes the paper.

\section{Related Work}\label{sec:relwork}

\subsection{Classical and Deep Learning Approaches}\label{subsec:classical}

Early work applied classical machine learning methods, such as Support Vector Machines (SVM),
Logistic Regression, and Naive Bayes, to text classification using TF-IDF or bag-of-words
features. Ahmad~et al.~\cite{ahmad2025} experimented with several classifiers, including SVM,
Random Forest, KNN, and Naive Bayes, on the Reuters-21578 dataset and reached accuracies of up
to 97\%, but did not explain how their models arrived at their predictions. Koszty\'an~et al.~\cite{kosztyan2025} showed that a tuned XGBoost classifier
can automatically label scientific papers by research methodology with 90--95\% average
accuracy, illustrating the practical value of automated document classification for
organisational review workflows. Among the earliest deep-learning systems for sensitive-document
classification is that of Alzhrani~et al.~\cite{alzhrani2019}, who split long paragraphs of
WikiLeaks cables into multiple sequences to focus on the parts most relevant to
classification. Hart~et al.~\cite{hart2011} and Saritha and Kumar~\cite{saritha2025} applied
text classification and deep-learning-based models, respectively, to DLP tasks, the latter
reporting 97.67\% accuracy, but neither addressed the heterogeneous organisational document
landscape considered here.

\subsection{Transformer-Based Models}\label{subsec:transformer}

BERT~\cite{devlin2019} encodes text bidirectionally and can be fine-tuned for a specific task
with relatively little labelled data. Petrolini~et al.~\cite{petrolini2022} fine-tuned BERT for
sensitive-data detection, achieving strong performance without explaining their decisions.
Minaee~et al.~\cite{minaee2021} reviewed more than 150 deep-learning text classification
models and found that transformer-based approaches consistently outperform earlier CNN- and
RNN-based architectures, while highlighting the need for testing in real organisational
settings. Large language models have also been surveyed as general-purpose text classifiers
\cite{minaee2025}; prompting or fine-tuning such models for sensitivity detection is, however,
considerably more expensive and harder to control than fine-tuning a compact encoder.
% TODO(author): the original sentence said minaee2025 "fine-tuned GPT-4 for zero-shot
% sensitive-text detection in emails", but minaee2025 in the bibliography is the survey
% "Large Language Models: A Survey". If you meant a different paper, cite it here instead.

\subsection{Document Layout and Explainability}\label{subsec:layout}

Huang~et al.~\cite{huang2022} presented a model that jointly encodes text and layout for
document understanding. % TODO(author): the original text reported "89% accuracy on contracts
% and invoices"; LayoutLMv3 reports results on RVL-CDIP, FUNSD, CORD, etc. Please verify the
% figure before restoring it.
McDonald~\cite{mcdonald2019} introduced a technology-assisted sensitivity review framework,
showing that AI-prioritised review improved human reviewers' accuracy and speed.
% TODO(author): the original text gave +38% accuracy and +72% speed; verify against the thesis.
Rudin~\cite{rudin2019} argued that models used for high-stakes decisions must be explainable,
and Gaspar~et al.~\cite{gaspar2024} showed that combining LIME and SHAP with security
classification systems improves analyst efficiency and decision transparency. Shapley-value
attribution also remains an active research area: Zhao~et al.~\cite{zhao2025shapg} proposed
ShapG, a graph-based Shapley-value method that reduces the cost of computing feature
importance.

\subsection{Multi-Agent and Long-Document Modelling}\label{subsec:mas}

Multi-agent designs built on large language models have recently been applied to engineering
problems. Ferraro et~al.~\cite{ferraro2025} proposed a hierarchical multi-agent framework for
insider threat detection in which specialised agents analyse categorised activity logs and a
supervisor agent issues the final decision. Like the present work, they prioritise recall over
precision to minimise missed threats, and their ablation shows that removing the supervisor
agent sharply degrades performance. Alevizos et~al.~\cite{alevizos2026} use a multi-agent
framework built on a fine-tuned large language model to generate sustainable, biomimetic
furniture designs. To our knowledge, however, no prior work has applied a multi-agent
architecture to document sensitivity classification (Table~\ref{tab:relwork}). Long-document strategies fall into three
broad families: sub-quadratic attention mechanisms, hierarchical chunk-then-aggregate
approaches, and multiple-instance-learning-style segment bags \cite{survey2025}. Tsirmpas
et~al.~\cite{tsirmpas2024} likewise identify input-length limits as a central obstacle for
neural classification and summarization of long texts, and review truncation, chunking, and
efficient-attention remedies;
hierarchy-aware architectures have also been explored for related structured tasks such as
hierarchical text categorization \cite{kumar2024}. Mixture-of-experts and gated ensembling
architectures \cite{jacobs1991,fedus2022} use learned, input-dependent weighting to combine
sub-models, a principle also reflected in dynamic BERT-based multichannel fusion for
classification \cite{zhou2023}; the Channel Critic Agent introduced here follows this
tradition but controls cross-boosting between two pretrained transformer encoders. Iterative
message passing has been studied extensively in graph neural networks \cite{li2016} and, more
recently, in multi-round LLM debate frameworks \cite{du2024}; the Iterative Consultation
mechanism developed here can be viewed as a two-node instance of this family. Adaptive
Computation Time \cite{graves2016} and confidence-based early exiting \cite{schuster2022}
motivate the adaptive consultation used here, in line with the broader emphasis on
computational efficiency in deployed machine learning systems \cite{dantas2024}.

\begin{table}[!htbp]
\small
\centering
\caption{Comparison of methodologies in prior sensitivity-classification studies. Scores are
reported on different datasets and are not directly comparable. n/r: not reported as a single
figure.}\label{tab:relwork}
\begin{tabular}{@{}lccccc@{}}
\toprule
Study & Method & Dataset & Score & XAI & MAS \\
\midrule
Alzhrani et al.~\cite{alzhrani2019} & CNN & PlusD & F1=0.91 & No & No \\
Petrolini et al.~\cite{petrolini2022} & BERT & Privacy & F1=0.95 & No & No \\
McDonald~\cite{mcdonald2019} & Classif. & Gov. recs. & BA=0.70 & No & No \\
Ahmad et al.~\cite{ahmad2025} & SVM/RF/NB & Reuters & Acc=0.97 & No & No \\
Hart et al.~\cite{hart2011} & Text clf. & DLP & n/r & No & No \\
\midrule
Zainab et al.~\cite{zainab2026benchmark} & BERT & Strategic 16K & F1=0.8933 & No & No \\
This work (IC-MAS) & IC-MAS & Strategic 16K & F1=0.9187 & Yes & Yes \\
\bottomrule
\end{tabular}
\end{table}
% Note: F1 for "This work" updated to the Full MAS (0.9187); use 0.9123 if you prefer the core model.

IC-MAS is, to our knowledge, the first architecture to integrate critic-controlled gated
fusion with an adaptive, iterative agent consultation mechanism for document sensitivity
classification. The next two sections lay its empirical groundwork: the leakage-controlled
Strategic~16K corpus (Section~\ref{sec:dataset}) and the single-encoder baselines it must
outperform (Section~\ref{sec:baseline}).

\section{Dataset and Leakage-Controlled Benchmark Corpus}\label{sec:dataset}

The corpus construction and leakage-removal protocol summarized in this section were first
described in our previous report \cite{zainab2026benchmark}, which gives full methodological
details and further validation experiments. We recap them here because this corpus, unchanged,
is used to train and evaluate the proposed architecture throughout this paper.

\subsection{Dataset Source}\label{subsec:source}

The data come from the WikiLeaks Public Library of US Diplomacy (PlusD), which contains 251,287
diplomatic cables, each labelled by government officials with one of seven classifications:
Unclassified, Confidential, Limited Official Use, Secret, Unclassified//For Official Use Only,
Confidential//Noforn, and Secret//Noforn. These are actual classification decisions rather than
crowd-sourced annotations~\cite{zainab2026benchmark}. The documents were retrieved through a
custom extraction pipeline that traversed the paginated PlusD interface,
extracting each document's content, original classification label, and canonical identifier;
HTML artefacts were stripped and records with missing or inconsistent metadata were discarded,
yielding over 100{,}000 deduplicated documents~\cite{zainab2026benchmark}.

\subsection{Corpus Construction: Strategic 16K}\label{subsec:corpus}

The seven original labels were mapped to a binary schema, Non-Sensitive (7{,}613 documents)
and Sensitive (8{,}387 documents), giving a total of 16{,}000 documents (52.4\% Sensitive,
47.6\% Non-Sensitive)~\cite{zainab2026benchmark}. The label mapping and the sampling procedure
used to draw these documents from the deduplicated collection are explained in
\cite{zainab2026benchmark}. Deduplication was performed before splitting, and stratified
5-fold sampling with a fixed random seed of 42 was used throughout for
reproducibility~\cite{zainab2026benchmark}. This binary schema and corpus are used, unchanged,
for every experiment in this paper.

\subsection{Leakage Removal Protocol}\label{subsec:leakage}

A pervasive but underreported problem when using PlusD as a training corpus is \textit{label
leakage}: residual classification markers embedded in document bodies that let a model exploit
surface shortcuts rather than content-based sensitivity signals~\cite{zainab2026benchmark}.
Three categories of residual artefacts were identified and removed: (i)~inline paragraph
markers such as (C), (S), (U), and (SBU); (ii)~repeated classification phrases in the body
(e.g., ``this cable is classified SECRET''); and (iii)~distribution-notice boilerplate (e.g.,
``Sensitive But Unclassified, Not for Internet Distribution''). Classification-related terms
occurring in natural sentence contexts (e.g., a cable discussing nuclear confidentiality
agreements) were deliberately left unmodified, since the goal was artefact removal rather than
content suppression~\cite{zainab2026benchmark}.

Two independent checks confirmed successful leakage removal. First, the top-15 TF-IDF features
per class on the cleaned corpus contained no classification-related tokens. Second,
Table~\ref{tab:leakage} shows that TF-IDF + Logistic Regression accuracy drops by nearly 13
points after leakage removal, indicating that the inflated raw-corpus performance was due to
residual artefacts rather than semantic understanding. A length-only baseline also achieved
substantially lower accuracy than every evaluated model, indicating that document length
alone is not an exploitable shortcut (see \cite{zainab2026benchmark} for the full result).

\begin{table}[!htbp]
\small
\centering
\caption{Impact of label leakage on TF-IDF + Logistic Regression performance}\label{tab:leakage}
\begin{tabular}{@{}lcc@{}}
\toprule
Dataset & Accuracy & F1 \\
\midrule
Raw PlusD (leaky) & 99.0\% & 99.0\% \\
Strategic 16K (cleaned) & 86.26\% & 86.83\% \\
\bottomrule
\end{tabular}
\end{table}

\subsection{Document Length Distribution}\label{subsec:length}

The median length of a sensitive document in Strategic~16K is 276 words, versus 64 words for
non-sensitive documents, and a non-trivial fraction of sensitive cables are longer than
1{,}400 words. This asymmetry matters for any transformer with a fixed input window: when the
same truncation is applied to every document, it is disproportionately likely to discard
evidence from the class that most needs to be detected correctly.
Section~\ref{sec:baseline} quantifies the resulting baseline performance, and
Section~\ref{sec:architecture} then develops the dual-window architecture that targets this
shortfall.

\subsection{Evidence Location Distribution}\label{subsec:evidencelocation}

A qualitative review of Strategic~16K cables during corpus preparation revealed a fairly
consistent structural pattern: sensitivity-relevant content (classification rationale,
substantive assessments, and concluding remarks) tends to cluster near the beginning and end
of a cable rather than being evenly distributed. Diplomatic cables typically open with a
subject line and framing paragraph that establish the topic and its sensitivity, and close
with a concluding assessment, comment, or classification justification; the intervening body
more often consists of descriptive or background material. This observation is qualitative
rather than a formally quantified corpus statistic, but it motivates the window pair used in
Section~\ref{sec:architecture}: reading the \emph{beginning and end} of a document, rather
than, for instance, two windows from the first half or a first window paired with a randomly
sampled middle segment. An ablation in Section~\ref{subsec:negative} tests this choice
against a first-window-doubled alternative and finds it consistently preferable, providing
model-level support for the observed pattern.

\section{Baseline Benchmark: Classical and Transformer-Based Models}\label{sec:baseline}

This section recaps the baseline results of \cite{zainab2026benchmark}, which serve as the
point of comparison for the architecture proposed in Section~\ref{sec:architecture}.

\subsection{Evaluation Protocol}\label{subsec:evalprotocol}

Six models were evaluated using 5-fold stratified cross-validation on Strategic~16K, with the
class ratio preserved across folds and a fixed random seed of 42~\cite{zainab2026benchmark}.
Weighted F1 and accuracy were the primary metrics; Sensitive-class recall was tracked as a
secondary criterion, since a missed sensitive document is operationally costlier than a false
alarm~\cite{zainab2026benchmark}. Three classical TF-IDF models (Logistic Regression, Linear
SVM, and Multinomial Naive Bayes, with hyperparameters tuned via RandomizedSearchCV) and three
transformer models (BERT \cite{devlin2019}, RoBERTa \cite{liu2019}, and ELECTRA
\cite{clark2020}) fine-tuned end-to-end with a maximum input length of 256 tokens were
compared.

\subsection{Results}\label{subsec:baselineresults}

Table~\ref{tab:baseline} presents the 5-fold cross-validation results. BERT leads with 89.33\%
F1 and 89.14\% accuracy, 2.38~pp F1 above the best classical model (SVM, 86.95\%~F1) and
4.41~pp above Naive Bayes. Paired $t$-tests confirmed that BERT's advantage over the classical
models was statistically significant (e.g., BERT vs.\ SVM: $t=15.19$, $p<0.001$).
% TODO(author): the original text said "every transformer-vs-classical comparison" was
% significant, which cannot hold for RoBERTa (it scores below LR and SVM). Restate precisely
% which comparisons were significant.
RoBERTa (86.51\%~F1) performs slightly below the best classical models, which we attribute to
its pretraining not transferring as well to PlusD's vocabulary and register. ELECTRA achieves
higher Sensitive Recall than BERT (87.24\% vs.\ 86.66\%) at lower precision, an early instance
of the recall/precision tension that motivates Sections~\ref{sec:architecture}--\ref{sec:results}.
Among classical models, SVM performs best using only CPU resources, making TF-IDF classifiers a
practical low-cost deployment option.

\begin{table}[!htbp]
\small
\centering
\caption{5-fold cross-validation results on Strategic 16K}\label{tab:baseline}
\begin{tabular}{@{}lcccc@{}}
\toprule
Model & Acc. & F1 & S-Recall & S-Prec. \\
\midrule
BERT & 89.14\% & 89.33\% & 86.66\% & 92.21\% \\
ELECTRA & 88.57\% & 88.90\% & 87.24\% & 90.70\% \\
RoBERTa & 85.85\% & 86.51\% & 86.57\% & 86.45\% \\
LR + TF-IDF & 86.26\% & 86.83\% & 86.41\% & 87.25\% \\
SVM & 86.35\% & 86.95\% & 86.74\% & 87.16\% \\
Na\"ive Bayes & 83.91\% & 84.92\% & 86.40\% & 83.48\% \\
\bottomrule
\end{tabular}
\end{table}

\subsection{Transition to the Proposed Architecture}\label{subsec:transition}

The central conclusion of this benchmark is that a single-encoder model has a structural
limitation for security-critical deployment: no evaluated model can see beyond its fixed
256-token input window. Section~\ref{sec:architecture} presents a window-based multi-agent
architecture that addresses this limitation directly.

\section{Proposed Architecture: IC-MAS}\label{sec:architecture}

\subsection{Motivation}\label{subsec:motivation}

\begin{figure}[!htbp]
\centering
\includegraphics[width=\textwidth]{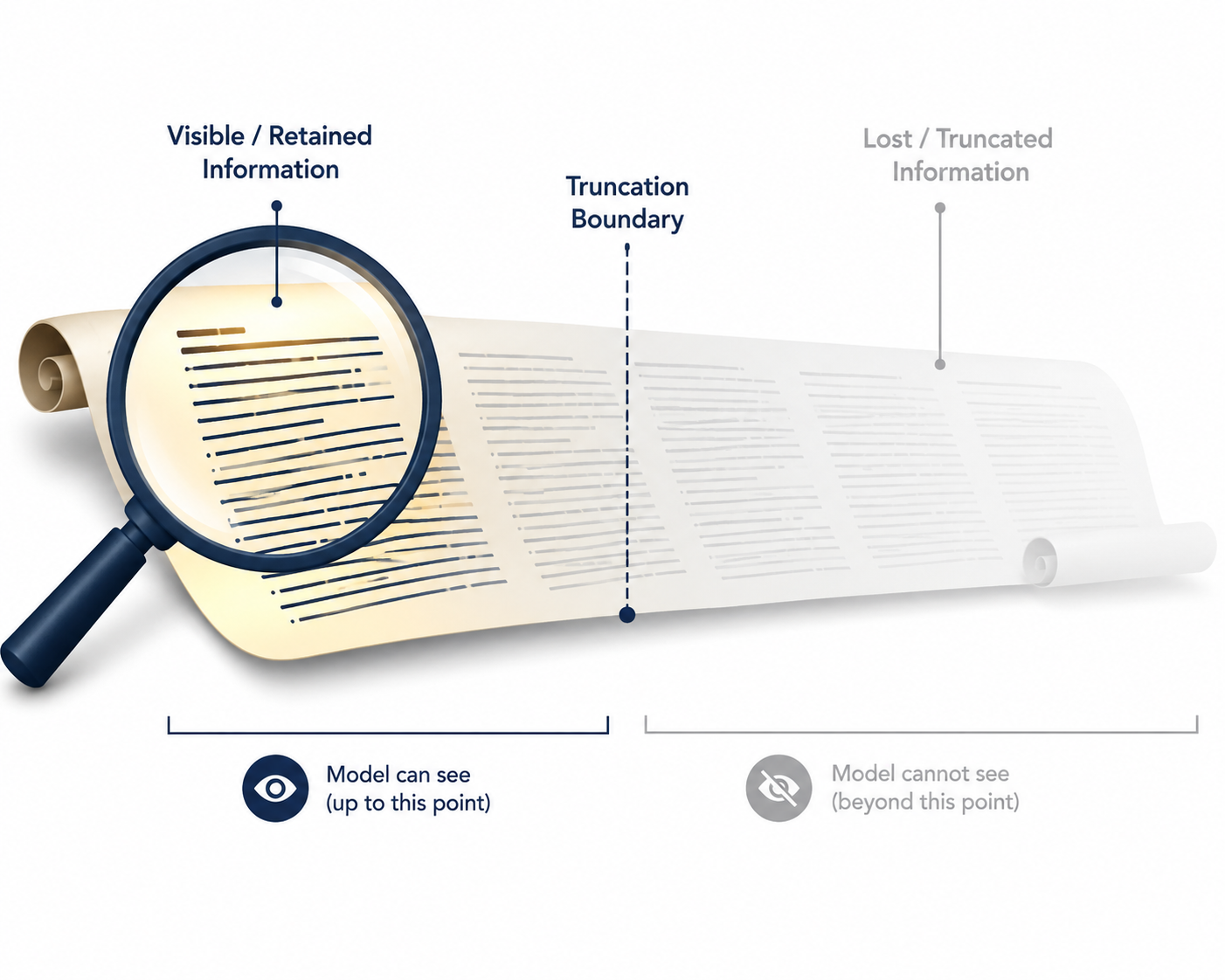}
\caption{Illustration of the truncation problem: a fixed-length input window retains only a
portion of the document, and content beyond the truncation boundary is never presented to the
model.}\label{fig:truncation}
\end{figure}

Given the length asymmetry documented in Section~\ref{sec:dataset}, the 256-token ceiling of
Section~\ref{sec:baseline} means that any classifier limited to a single fixed window is
systematically disadvantaged on the class that most needs correct detection
(Fig.~\ref{fig:truncation}): evidence beyond the window boundary is never visible to the model,
however task-relevant it is. IC-MAS addresses this by reading two windows, the beginning and
the end of the document, and reconciling them through learned agent interaction. On the first
window, a Channel Critic Agent (Section~\ref{sec:critic}) controls the fusion of two encoders
through document-specific trust weights that are learned purely from the downstream
classification loss rather than fixed at design time.

\subsection{Why Not a Long-Context Transformer}\label{sec:why-not-longcontext}

A natural question is why we do not process the entire document with a long-context
transformer (e.g., Longformer, BigBird), which removes truncation by extending the attention
window, rather than reading two fixed windows and reconciling them. Three considerations
motivated the window-and-reconcile design.

First, a long-context transformer's cost grows with document length, even though, in this
corpus, the decision does not require reasoning over every intervening token, only over the
regions that carry sensitivity-relevant evidence. The cost of our design is bounded per
document: four 256-token encoder passes (BERT on each window, RoBERTa and the frozen
DistilBERT critic on the first window) followed by a small number of consultation rounds over
768-dimensional vectors. Adaptive halting (Section~\ref{sec:adaptive}) further reduces the
number of consultation rounds for straightforward documents, although the encoder passes,
which dominate the total cost, are unaffected.

Second, the architecture reconciles evidence in a compact, task-specific representation space
rather than in token space. Each window is first compressed into a single 768-dimensional
belief vector ($\mathbf{v}_\mathrm{start}$, $\mathbf{v}_\mathrm{end}$) before any cross-window
interaction; the Consultation Agents then refine these two vectors against one another
(Eqs.~\ref{eq:msg}--\ref{eq:gruB}) without re-examining the tokens. This late-fusion,
iterative-refinement design contrasts with the joint attention over all token pairs performed
by a long-context transformer, which may dilute the signal of a small evidence-bearing region
within a single document-level representation.

Third, the design is falsifiable against a natural extension of itself: the $N$-chunk ablation
(Section~\ref{subsec:negative}) evaluated whether extending consultation to four regions,
covering more of the document, improves on two windows. It did not; recall dropped by 2.06~pp.
For this corpus, the two-window split appears to capture the regions carrying most
sensitivity-relevant evidence, and broader coverage does not help unless the aggregation
mechanism is suited to it. This ablation does not, however, substitute for a direct comparison
with long-context transformers, which we discuss as a limitation in
Section~\ref{sec:limitations}.

\subsection{System Overview}\label{subsec:overview}

Throughout this paper, an \emph{agent} denotes a specialized computational component with a
well-defined input, output, and role in the pipeline. The four core classification agents
(Channel Critic, Dual-Window Reader, Consultation, and Consensus Agents) are trainable modules:
fine-tuned transformer encoders (BERT, RoBERTa), lightweight feed-forward or recurrent modules
(the Channel Critic Agent's MLP, the Consultation Agents' GRU cells), and none of them prompts a
language model or reasons via free-text generation. Domain Evidence Agent combines lexical and embedding-based scoring. Only the
optional Decision \& Verification Agent (Section~\ref{sec:decision}) and the post-hoc
explanation stage (Section~\ref{sec:xai}) use a small instruction-tuned language model, and in
both cases as a reasoning tool inside a fixed control structure. The multi-agent framing thus
follows the classical, modular-systems sense of the term: multiple specialized components, each
with a narrow competency, coordinated to produce a decision that no single component produces
alone.

The agents communicate exclusively through a shared \texttt{Blackboard} object
\cite{hayesroth1985}. Given a document $D$ whose token sequence exceeds the encoder's maximum
input length $L=256$, the model defines two fixed-length windows, the first window
$D_\mathrm{first}$ and the last window $D_\mathrm{last}$, motivated by the evidence-location
pattern of Section~\ref{subsec:evidencelocation}: a first-and-last pair is expected to retain
more decision-relevant evidence per token than alternative pairings.
% TODO(author): state how documents with at most L=256 tokens are handled (e.g., is
% D_last = D_first?), and whether the two windows overlap for documents of 257--512 tokens.
The task is binary classification, $y \in \{\mathrm{sensitive}, \mathrm{non\text{-}sensitive}\}$,
using evidence from both windows without any hand-crafted combination rule.

\begin{figure}[!htbp]
\centering
\includegraphics[width=\textwidth]{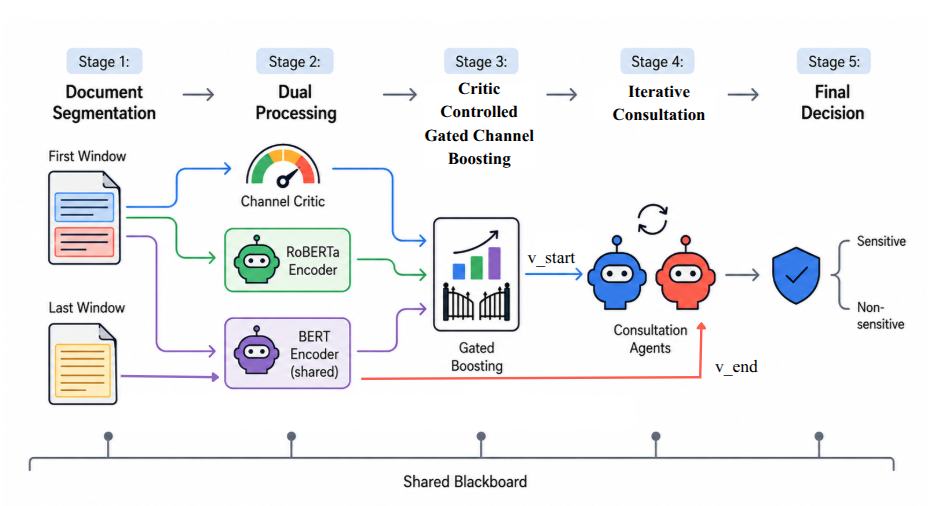}
\caption{High-level overview of the IC-MAS pipeline: document segmentation, asymmetric
dual-window processing, iterative consultation, adaptive halting, and final decision, all
coordinated through the shared Blackboard.}\label{fig:pipeline}
\end{figure}

\begin{figure}[!htbp]
\centering
\includegraphics[width=0.92\textwidth]{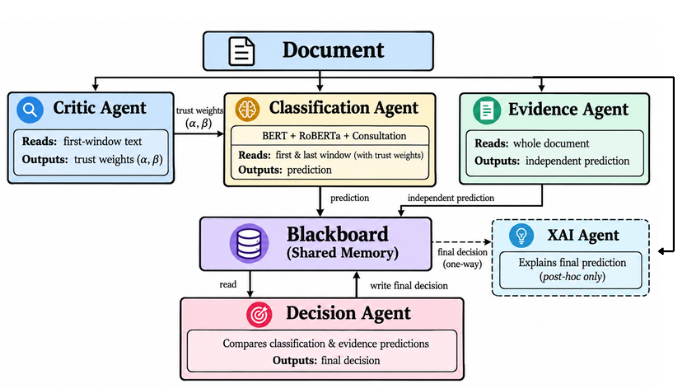}
\caption{Overall IC-MAS
architecture with the Domain Evidence and Decision \& Verification Agents. The Classification
Agent comprises the Dual-Window Reader, Consultation, and Consensus Agents.}\label{fig:schematic}
\end{figure}
Figure~\ref{fig:pipeline} gives a high-level view of the pipeline; Fig.~\ref{fig:schematic}
presents the full technical schematic, including the keys exchanged through the shared
Blackboard. Throughout, \emph{Classification Agent}
refers collectively to the Dual-Window Reader, Consultation, and Consensus Agents.

\subsection{Channel Critic Agent}\label{sec:critic}

\begin{figure}[!htbp]
\centering
\includegraphics[width=0.65\textwidth]{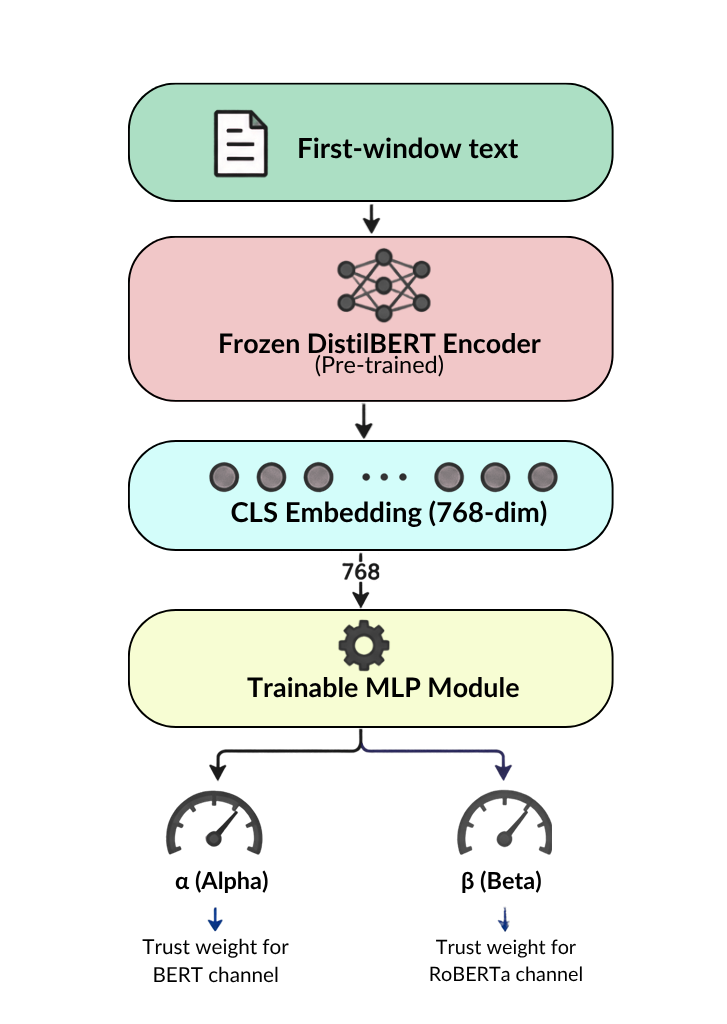}
\caption{Channel Critic Agent data flow: first-window text is encoded by a frozen DistilBERT,
reduced to its [CLS] embedding, and passed through a trainable MLP to produce the
document-adaptive trust weights $\alpha$ and $\beta$.}\label{fig:critic}
\end{figure}

The Channel Critic Agent judges, per document, how much trust to place in each of two
evidence sources about the first window, producing two trust weights that govern how strongly
each source influences the fused representation. The two sources are the [CLS] embeddings
$\mathbf{p}_1, \mathbf{p}_2 \in \mathbb{R}^{768}$ obtained by encoding the first window with
BERT and RoBERTa, respectively. A frozen DistilBERT~\cite{sanh2019} encoder followed by a
lightweight trainable MLP reads the same first-window text and produces the trust weights
(Fig.~\ref{fig:critic}). The MLP maps the DistilBERT [CLS] embedding to a 2-dimensional vector
of logits, to which a rescaled sigmoid,
\begin{equation}
\sigma_{[0.5,1.5]}(x) = 0.5 + \sigma(x),
\end{equation}
is applied elementwise:
\begin{equation}
(\alpha, \beta) = \sigma_{[0.5,1.5]}\big(\mathrm{MLP}(\mathrm{DistilBERT}(D_\mathrm{first})_{[\mathrm{CLS}]})\big),
\end{equation}
where $\sigma$ is the logistic sigmoid, so that $\alpha,\beta \in [0.5,1.5]$. The Channel
Critic Agent receives no direct supervision on the ``correct'' trust weights; $\alpha$ and
$\beta$ are learned solely through gradients backpropagated from the classification loss.

\subsection{Dual-Window Reader Agent}\label{subsec:dualwindow}

The Dual-Window Reader Agent reads both windows under a deliberately asymmetric design. It
reads the first window with both BERT and RoBERTa, cross-boosted using the Channel Critic
Agent's trust weights (Section~\ref{sec:critic}), and the last window with BERT alone
(Section~\ref{subsec:lastwindow}); the shared BERT encoder already receives gradient signal
from both windows during training, without a second dedicated encoder. The following
subsections describe the gated boosting step, the fusion step, and the last window's simpler
treatment.

\subsubsection{Gated Channel Boosting}\label{sec:gcb}

\begin{figure}[!htbp]
\centering
\includegraphics[width=0.75\textwidth]{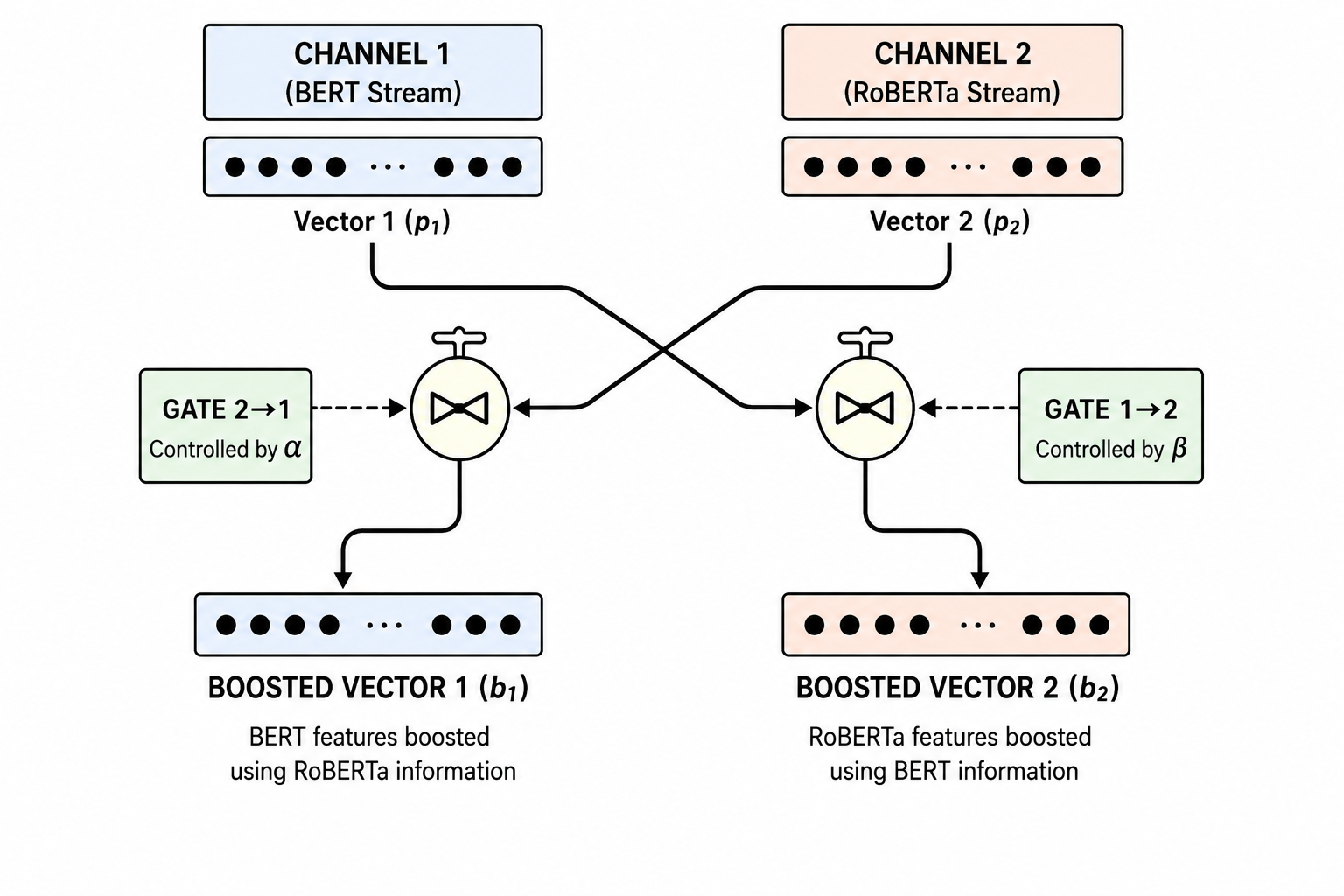}
\caption{Gated Channel Boosting: the BERT vector $\mathbf{p}_1$ and RoBERTa vector
$\mathbf{p}_2$ cross-gate one another, modulated by the critic's trust weights $\alpha$ and
$\beta$, producing boosted vectors $\mathbf{b}_1$ and $\mathbf{b}_2$.}\label{fig:boosting}
\end{figure}

The trust weights modulate a gated cross-boosting step that lets each encoder's representation
be selectively amplified by signal from the other:
\begin{align}
\mathbf{b}_1 &= \mathrm{LN}\big(\mathbf{p}_1 + \mathbf{p}_1 \odot \sigma(W_{21}\mathbf{p}_2)\cdot\alpha\big), \\
\mathbf{b}_2 &= \mathrm{LN}\big(\mathbf{p}_2 + \mathbf{p}_2 \odot \sigma(W_{12}\mathbf{p}_1)\cdot\beta\big),
\end{align}
where $\mathrm{LN}(\cdot)$ denotes Layer Normalization, $W_{21}, W_{12} \in
\mathbb{R}^{768\times768}$ are learned linear projections, $\sigma(W_{21}\mathbf{p}_2)$ and
$\sigma(W_{12}\mathbf{p}_1)$ are learned per-dimension gates, and $\odot$ denotes elementwise
multiplication. This mechanism (Fig.~\ref{fig:boosting}) is referred to as Channel Boosting
(CB).

\subsubsection{Fusion}\label{subsec:fusion}

The boosted representations $\mathbf{b}_1,\mathbf{b}_2$ are combined into a single vector
$\mathbf{v}_\mathrm{start}$ before consultation with the last window. Two strategies were
evaluated: a 2-layer Transformer self-attention fusion (the initial design, following the
multi-attention transformer-encoder fusion used in multimodal representation learning
\cite{liu2024}), and elementwise max-pooling,
\begin{equation}
\mathbf{v}_\mathrm{start} = \max(\mathbf{b}_1, \mathbf{b}_2).
\end{equation}
Section~\ref{sec:fusionablation} shows that max-pooling matches or exceeds Transformer fusion
with no additional parameters, so it is adopted in the final design.

\subsubsection{Last-Window Reading}\label{subsec:lastwindow}

The last window is encoded by the \emph{same} BERT instance used for the first window (shared
weights, no additional parameters), $\mathbf{v}_\mathrm{end} =
\mathrm{BERT}(D_\mathrm{last})_{[\mathrm{CLS}]}$, providing the second of the two
positionally distinct evidence sources that the Consultation Agents
(Section~\ref{sec:iterconsult}) reconcile. A symmetric variant that gives the last window the
same critic-and-boosting treatment as the first showed no consistent improvement, supporting
the asymmetric design.
% TODO(author): if available, report the symmetric variant's numbers (a short table row or
% a sentence with accuracy/F1) instead of only "no consistent improvement".

\subsection{Consultation Agents}\label{sec:iterconsult}

\begin{figure}[!htbp]
\centering
\includegraphics[width=\textwidth]{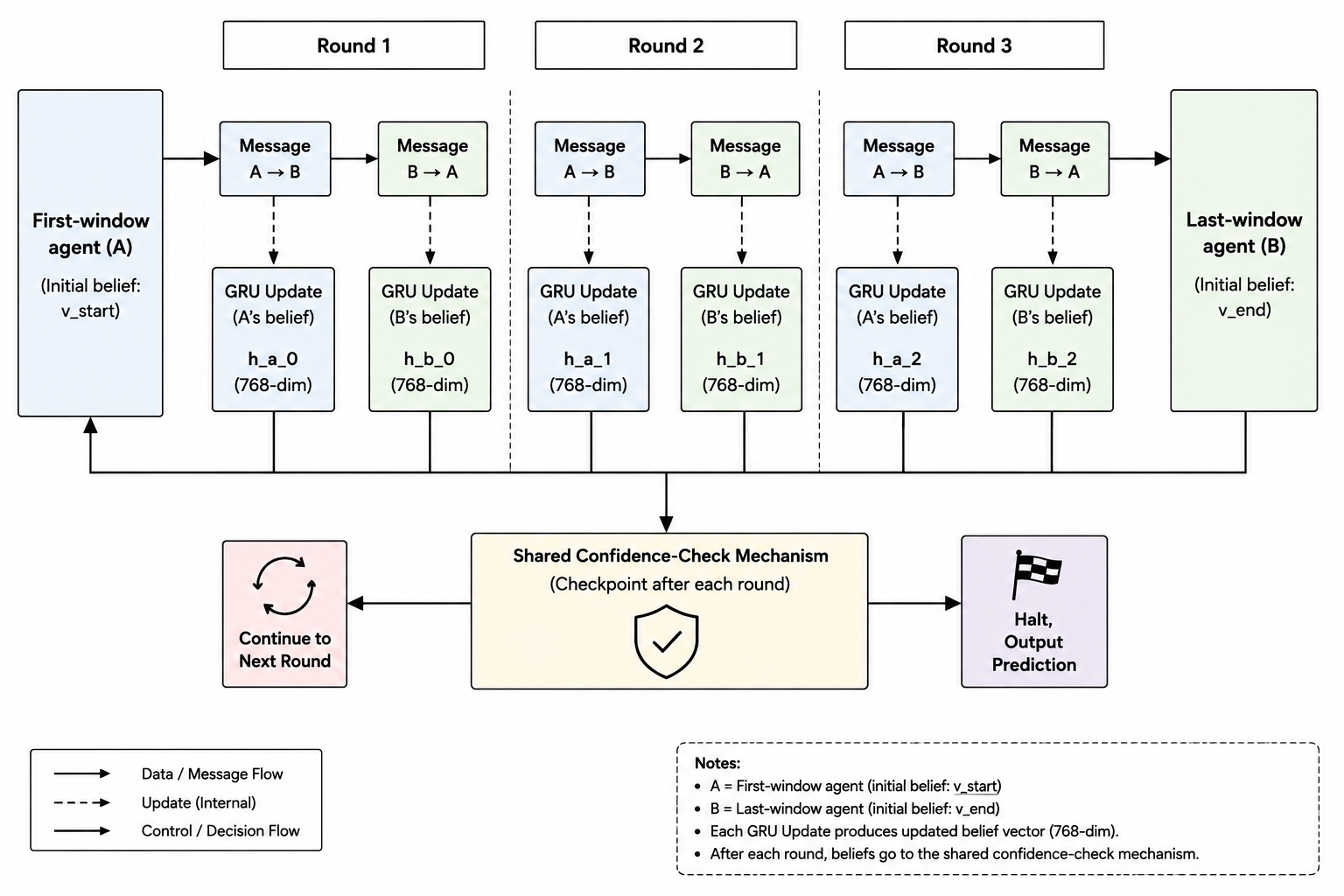}
\caption{Iterative consultation and adaptive halting. Two agents exchange messages over up to
three rounds (top); a shared confidence check on the combined output after each round decides
whether to continue or to halt and output the current prediction (bottom).}\label{fig:consultation}
\end{figure}

The Consultation Agents reconcile the two evidence sources produced by the Dual-Window Reader
Agent through iterative exchange rather than one-shot combination, so that evidence in one
window can influence the final decision even though the other window's encoder never observed
it. The architecture treats $\mathbf{h}_A^{(0)} = \mathbf{v}_\mathrm{start}$ and
$\mathbf{h}_B^{(0)} = \mathbf{v}_\mathrm{end}$ as the initial belief states of two agents,
which exchange messages over $T$ rounds ($T=3$ in the fixed-round configuration). At round $t$:
\begin{align}
\mathbf{m}_{B\to A}^{(t)} &= W_B \mathbf{h}_B^{(t-1)}, \quad
\mathbf{m}_{A\to B}^{(t)} = W_A \mathbf{h}_A^{(t-1)}, \label{eq:msg} \\
\mathbf{h}_A^{(t)} &= \mathrm{GRUCell}_A^{(t)}\big(\mathbf{m}_{B\to A}^{(t)}, \mathbf{h}_A^{(t-1)}\big), \label{eq:gruA} \\
\mathbf{h}_B^{(t)} &= \mathrm{GRUCell}_B^{(t)}\big(\mathbf{m}_{A\to B}^{(t)}, \mathbf{h}_B^{(t-1)}\big). \label{eq:gruB}
\end{align}
Here $W_A, W_B \in \mathbb{R}^{768\times768}$ are learned message projections, and each
GRU cell takes the incoming message as input
and the agent's prior belief as its hidden state. GRU cells are not shared between the two
agents or across rounds: each round $t \in \{1,\dots,T\}$ has its own pair of cells, giving
$2T$ independently parameterized GRU cells in total. Each GRU cell \cite{cho2014} learns, per
document and per round, how much of the incoming message to incorporate and how much of the
prior belief to retain, a soft, learned form of evidence aggregation in contrast to a
hand-crafted rule such as logical OR. Fig.~\ref{fig:consultation} illustrates this process
together with the adaptive halting mechanism (Section~\ref{sec:adaptive}).

\subsection{Consensus Agent}\label{subsec:consensus}

The Consensus Agent produces a single classification decision that depends jointly on both
Consultation Agents' final beliefs. After $T$ rounds, it computes the final logits from the
concatenation of both belief states,
\begin{equation}
\hat{y} = \mathrm{MLP}_\mathrm{cons}\big([\mathbf{h}_A^{(T)}; \mathbf{h}_B^{(T)}]\big),
\end{equation}
using a narrowing bottleneck ($1536 \to 256 \to 64 \to 2$).

\subsection{Blackboard-Style Adaptive Consultation}\label{sec:adaptive}

The number of consultation rounds adapts per document. A shared consensus head is applied
after every round, and a document halts once its confidence,
\begin{equation}
c^{(t)} = \big|\max(\mathrm{softmax}(\hat{y}^{(t)})) - 0.5\big| \times 2,
\end{equation}
exceeds a threshold $\tau=0.85$, following the Adaptive Computation Time paradigm
\cite{graves2016}.
% TODO(author): state how tau was chosen (and on which data), and how halting is handled
% during training (e.g., are all rounds supervised?).
Section~\ref{sec:adaptiveeff} shows that this reduces the average number of consultation
rounds by 54\% while matching or improving fixed-round performance.

\subsection{Domain Evidence Agent}\label{sec:evidence}

The Domain Evidence Agent independently judges a document's sensitivity from its
\emph{entire} content, providing a second opinion that the Decision \& Verification Agent can
consult when the Classification Agent's windowed view may have missed something. It works in
two stages. A lexical stage, fit once per fold on the training data only, assigns each
retained TF-IDF term a discriminative weight equal to the difference between its mean value on
sensitive and non-sensitive documents,
\begin{equation}
w(\mathrm{term}) = \bar{x}_\mathrm{sensitive}(\mathrm{term}) - \bar{x}_\mathrm{non\text{-}sensitive}(\mathrm{term}),
\end{equation}
and a document's lexical evidence score is the length-normalized sum of its matched terms'
weights,
\begin{equation}
s_\mathrm{lex}(D) = \frac{1}{\sqrt{n}}\sum_{\mathrm{term}\in D} w(\mathrm{term}),
\end{equation}
where $n$ is the document's word count; the region (first, middle, or last) with the largest
total weight is recorded as the document's evidence location. A semantic stage supplements
this exact-match signal using a separate, domain-adapted DistilBERT encoder (distinct from the
frozen DistilBERT instance inside the Channel Critic Agent), comparing the document's embedding
with per-class reference embeddings via cosine similarity, $s_\mathrm{sem}(D)$.
% TODO(author): add one clause on how this DistilBERT was domain-adapted (e.g., fine-tuned on
% each fold's training split only), so readers can confirm no test-fold data was used.
When the lexical stage flags the middle region, this comparison is computed on that region's
text, so the semantic signal can also see evidence outside the Classification Agent's two
windows. The two stages are blended, $s(D) = 0.7\,s_\mathrm{lex}(D) + 0.3\,s_\mathrm{sem}(D)$,
and the agent predicts \textsc{sensitive} if $s(D)>0$. Used as a standalone classifier under
the same 5-fold protocol, this agent reaches 76.03\% accuracy and 93.16\% Sensitive Recall,
higher recall than the core Classification Agent, but only 70.55\% Sensitive Precision: a
lexical and embedding-similarity scorer over the full document rarely misses sensitive
content, but without learned contextual reasoning it also flags many non-sensitive documents
that merely contain domain vocabulary. The Decision \& Verification Agent therefore treats
this agent's output as one input to a verification step rather than as a replacement
classifier.

\subsection{Decision \& Verification Agent}\label{sec:decision}

The Decision \& Verification Agent reconciles disagreements between the Classification
Agent's windowed prediction and the Domain Evidence Agent's full-document prediction, rather
than letting either win by default. It uses a lightweight language model
(Phi-4-mini-reasoning) as its reasoning tool within an explicit
Observe~$\to$~Reason~$\to$~Decide~$\to$~Act~$\to$~Feedback cycle: it \emph{observes} the
current Blackboard state (the Classification and Domain Evidence Agents' predictions),
\emph{reasons} about any disagreement using the language model, \emph{decides} on one of three
actions (\textsc{Accept}, \textsc{Investigate}, or \textsc{Escalate}), \emph{acts} by writing
its decision to the Blackboard, and, when it investigates, receives \emph{feedback} in the form
of the Domain Evidence Agent's recorded evidence location and text snippet before reasoning
again. A document-length gate restricts invocation to documents whose two windows do not
already cover the full text (more than 512 tokens), so the agent is invoked selectively rather
than on every document. As with the Channel Critic Agent, the underlying model is a tool the
agent uses; the agent itself is the surrounding control structure.
% TODO(author): state how ESCALATE is resolved when computing the reported metrics.

With every agent defined, Section~\ref{sec:methodology} specifies the experimental protocol,
and Section~\ref{sec:results} reports the results and ablation studies.

\section{Experimental Methodology}\label{sec:methodology}

\subsection{Evaluation Protocol and Metrics}\label{subsec:protocolmetrics}

All models are evaluated using 5-fold stratified cross-validation on Strategic~16K, with a
fixed random seed of 42 ensuring identical fold assignment across experiments. Four metrics
are reported: accuracy, F1-score, Sensitive-class recall, and Sensitive-class precision.
Sensitive recall is treated as the most important secondary metric, since in a
security-sensitive task a false negative costs substantially more than a false positive.

\subsection{Training Configuration}\label{subsec:trainconfig}

Unless otherwise noted, all models are trained for 5 epochs using AdamW with a learning rate
of $2\times10^{-5}$, weight decay of 0.01, a linear warmup ratio of 0.1, and a maximum sequence
length of 256 tokens per window.
% TODO(author): add batch size, whether the newly initialized modules (MLPs, GRUs, gates) use
% the same learning rate as the encoders, and whether a validation split inside each training
% fold was used for model/epoch/threshold selection. Reviewers will check for test-fold tuning.
Experiments were run on Kaggle T4 GPUs and a local DGX Station (4$\times$ NVIDIA V100 GPUs).

\subsection{Loss Function}\label{subsec:lossfn}

Several early experiments used class-weighted cross-entropy (sensitive class weighted
1.6$\times$) to target recall directly. To avoid confounding architecture with loss
configuration, Section~\ref{sec:classweight} reports an ablation that trains both baselines
and the proposed model under both loss configurations. The core architecture reported as the
headline result uses unweighted cross-entropy with label smoothing of 0.1.

\section{Results and Ablation Studies}\label{sec:results}

\subsection{Baseline and Core Architecture Comparison}\label{subsec:corecomparison}

Table~\ref{tab:main} presents the central comparison of this paper: single-encoder baselines,
the naive OR-logic combination heuristic, and the IC-MAS configurations, all trained with the
same unweighted objective.

\begin{table}[!htbp]
\small
\centering
\caption{Five-fold cross-validation results on Strategic 16K}\label{tab:main}
\begin{tabular}{@{}lcccc@{}}
\toprule
Model & Acc. & F1 & S-Recall & S-Prec. \\
\midrule
BERT (baseline) & 89.14\% & 89.33\% & 86.66\% & 92.21\% \\
RoBERTa & 85.85\% & 86.51\% & 86.57\% & 86.45\% \\
BERT+RoBERTa CB (Trans.) & 89.86\% & 90.44\% & 91.52\% & 89.39\% \\
CB + Dual-Window (OR) & 88.97\% & 89.83\% & 92.93\% & 86.93\% \\
CB + IC-MAS (Trans., fixed) & 90.57\% & 91.08\% & 91.81\% & 90.36\% \\
CB + IC-MAS (Trans., adapt.) & 90.40\% & 90.95\% & 92.07\% & 89.86\% \\
CB + IC-MAS (Max-Pool, fixed) & 90.66\% & 91.11\% & 91.33\% & 90.90\% \\
\textbf{CB + IC-MAS (Max-Pool, adapt.)} & \textbf{90.72\%} & \textbf{91.23\%} & 92.01\% & 90.46\% \\
CB + N-Chunk (N=4) & 89.26\% & 89.75\% & 89.75\% & 89.76\% \\
\bottomrule
\end{tabular}
\end{table}

Channel-Boosted fusion (row~3) yields a 4.86~pp gain in Sensitive Recall and a 1.11~pp gain in
F1 over the strongest single-encoder baseline. Naive OR-logic combination (row~4) raises recall
further, to 92.93\%, but at a 2.46~pp precision drop relative to CB alone, the precision
collapse that motivated the consultation-based approach (Section~\ref{sec:whyconsult}).
Replacing the OR rule with Iterative Consultation (rows~5--8) avoids this collapse in every
configuration while keeping recall above, or within 0.2~pp of, the CB-only level. Max-Pool
fusion with Adaptive Consultation (row~8) achieves the highest accuracy and F1 of any model in
the table, with only 1.38 average consultation rounds (Section~\ref{sec:adaptiveeff}).
Relative to the BERT-only baseline, this core configuration improves accuracy by 1.58~pp, F1 by
1.90~pp, and Sensitive Recall by 5.35~pp, at a cost of 1.75~pp precision, with no loss-function
intervention. Results are stable across folds (accuracy $90.72\%\pm0.19$~pp).

\begin{figure}[!htbp]
\centering
\includegraphics[width=\textwidth]{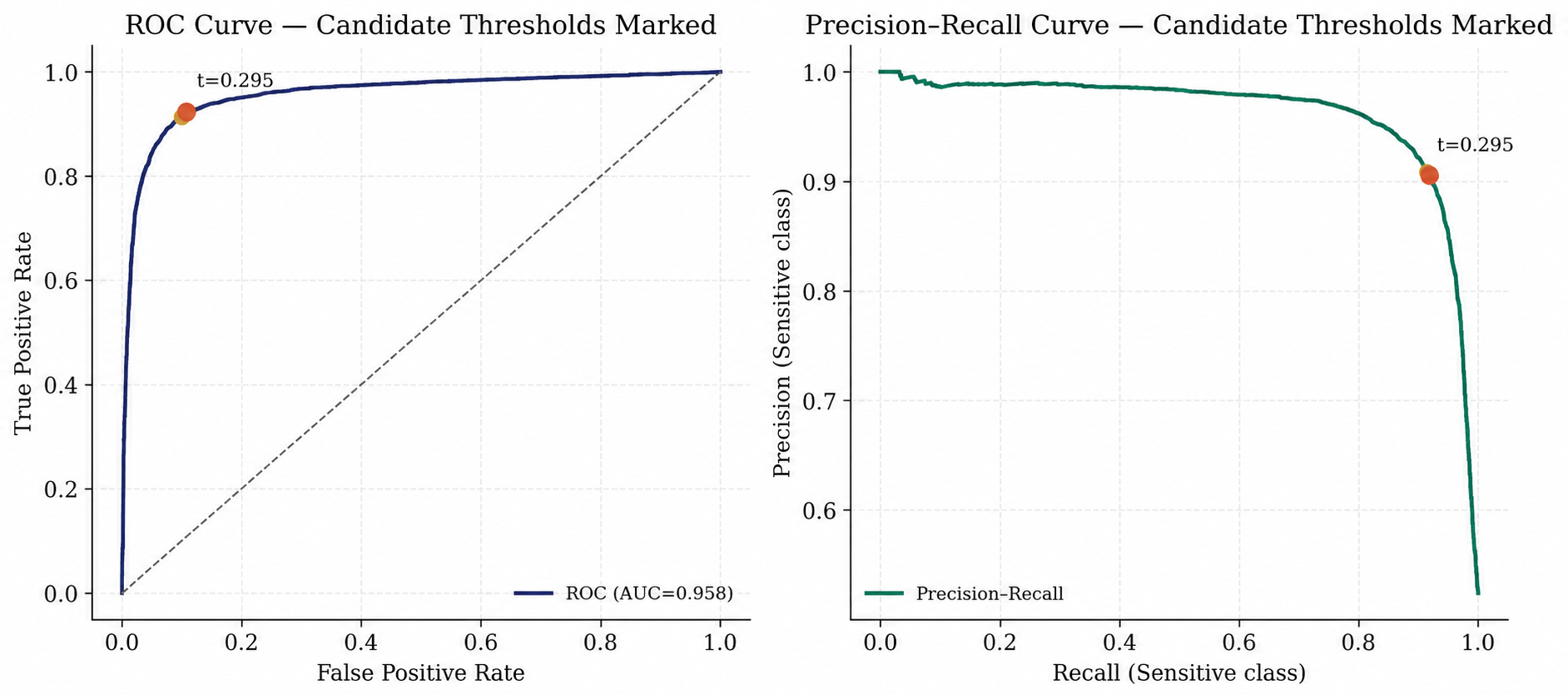}
\caption{ROC curve (left, AUC~$=0.958$) and precision--recall curve (right, Sensitive class)
for the core model, pooled across the 5 test folds. The marked point shows the operating
threshold $t=0.295$, chosen to reach 92\% Sensitive Recall.}\label{fig:roc}
\end{figure}

On the pooled test predictions ($n=16{,}000$), the core model achieves an ROC AUC of 0.958 and
an average precision of 0.962 (Fig.~\ref{fig:roc}).

% TODO(author): state on which data t = 0.295 was tuned. If it was tuned on the pooled test
% predictions, say so explicitly and present it as illustrative, not as a deployable setting.

\begin{figure}[!htbp]
\centering
\includegraphics[width=0.85\textwidth]{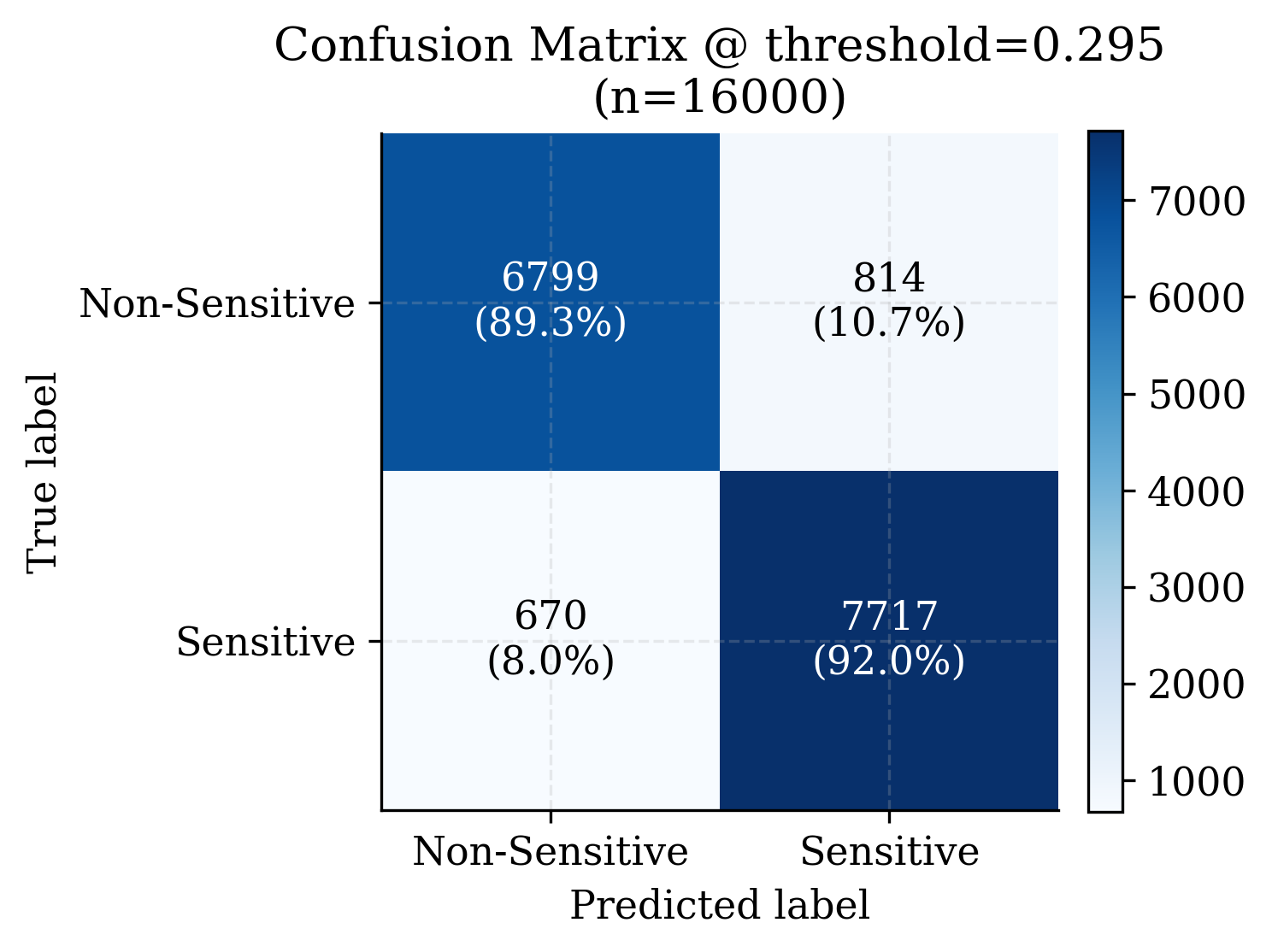}
\caption{Confusion matrix for the core model at the recall-targeted threshold ($t=0.295$),
pooled across the 5 test folds ($n=16{,}000$).}\label{fig:confmat}
\end{figure}

\begin{figure}[!htbp]
\centering
\includegraphics[width=0.68\textwidth]{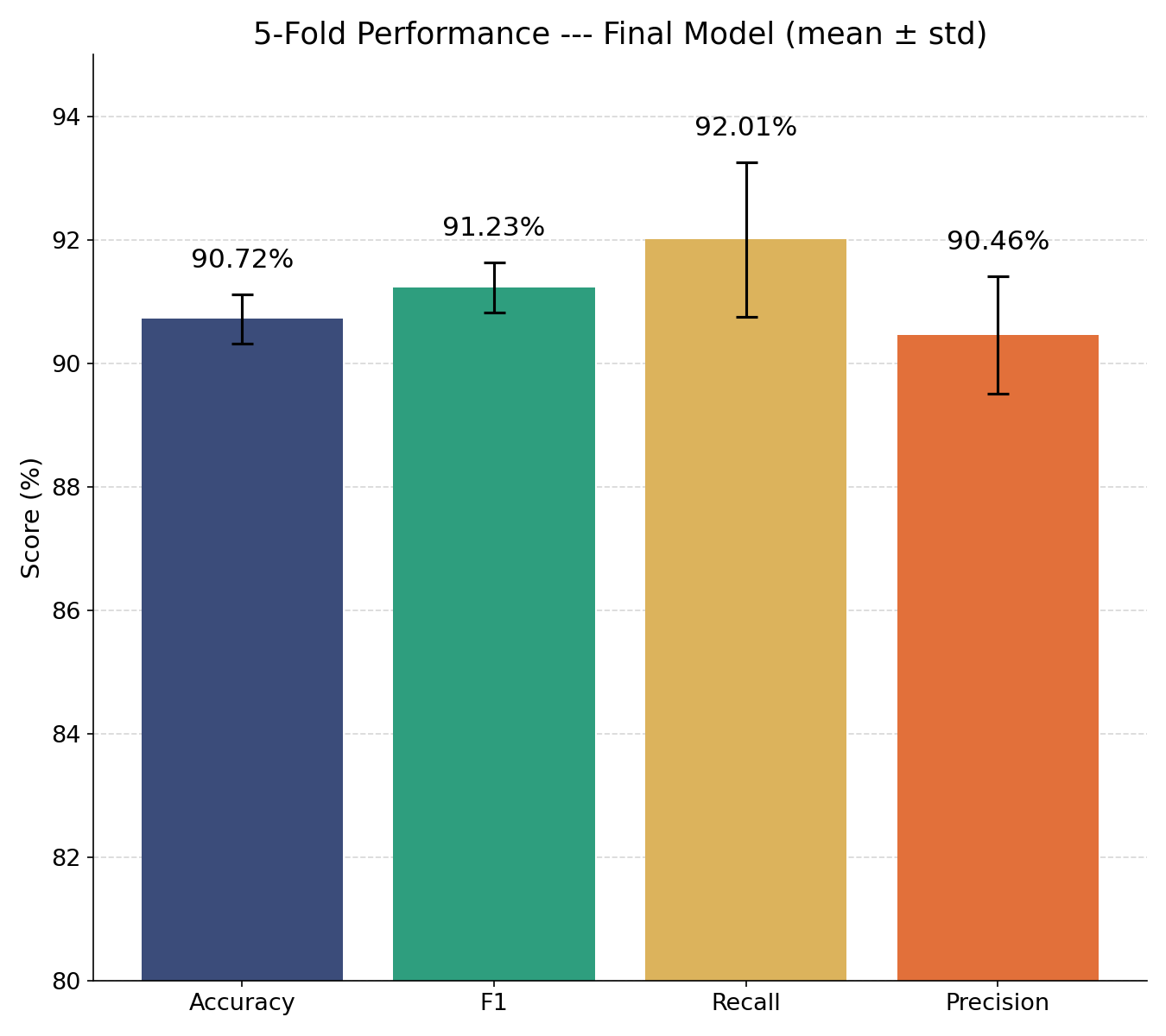}
\caption{Per-metric mean and standard deviation across the 5 cross-validation folds for the
core model, corresponding to the summary statistics in Table~\ref{tab:main}.}\label{fig:perfold}
\end{figure}

\begin{figure}[!htbp]
\centering
\includegraphics[width=\textwidth]{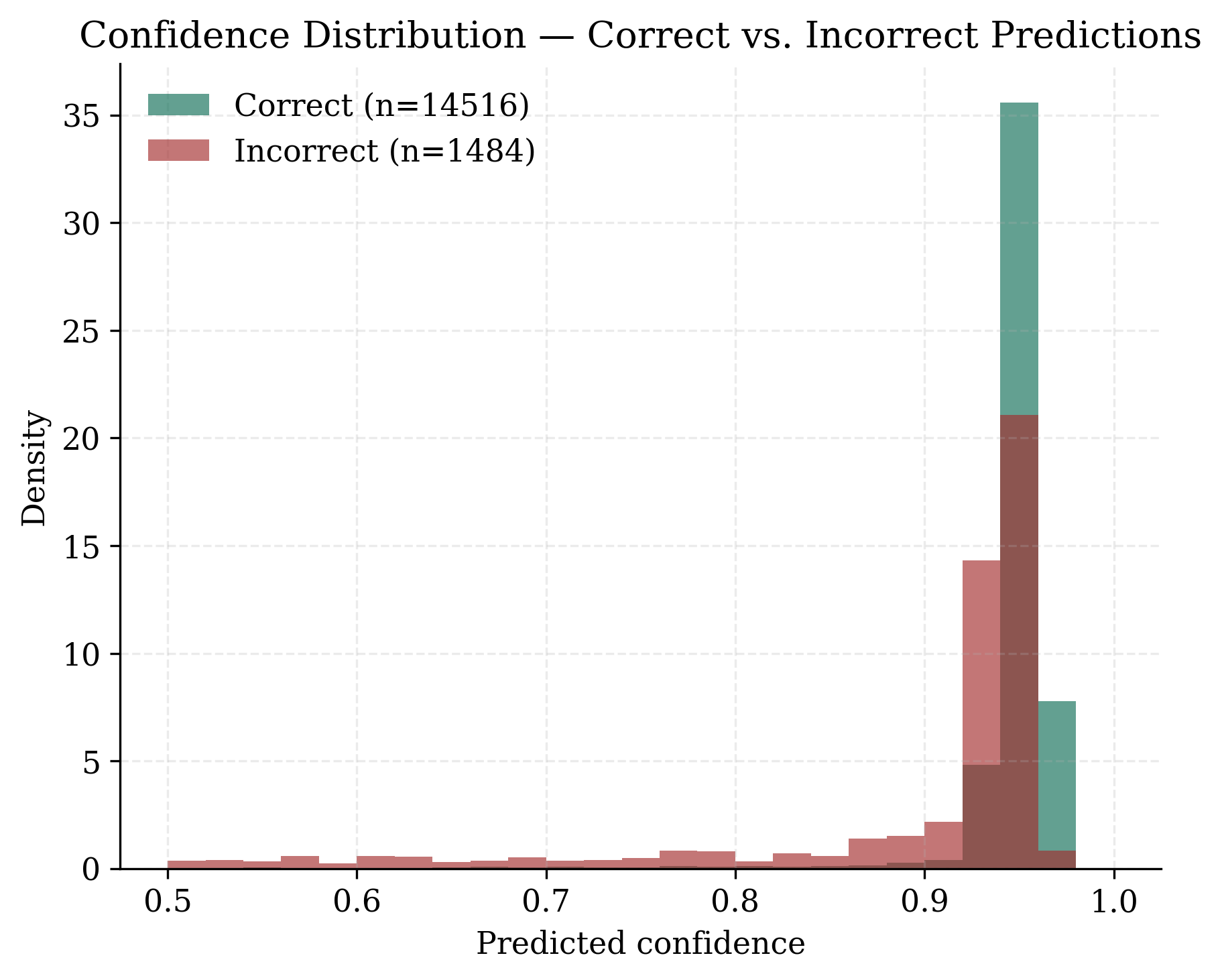}
\caption{Distribution of predicted confidence for correctly and incorrectly classified
documents at the default decision threshold ($t=0.5$).}\label{fig:confdist}
\end{figure}

Figures~\ref{fig:confmat} and~\ref{fig:perfold} show the pooled and per-fold results. The
confidence distribution (Fig.~\ref{fig:confdist}), computed at the default threshold $t=0.5$,
reveals a limitation discussed in Section~\ref{sec:limitations}: although only 1{,}484 of
16{,}000 predictions are incorrect at that threshold, correct and incorrect predictions are
both concentrated at high confidence, overlapping substantially in the 0.9--0.98 range.

\subsection{Fusion-Strategy Isolation Ablation}\label{sec:fusionablation}

Four strategies for combining $\mathbf{b}_1$ and $\mathbf{b}_2$ (concatenation, averaging,
max-pooling, and 2-layer Transformer self-attention) were evaluated with the Channel Critic
Agent and Gated Boosting held fixed, and differed only marginally in accuracy: max-pooling was
the strongest (90.10\%, Table~\ref{tab:rawfusion}), 0.24~pp above Transformer self-attention
fusion (89.86\%, Table~\ref{tab:main}), and adds no trainable parameters, so it was adopted.
% TODO(author): optional -- add the concatenation and averaging accuracies (one sentence or a
% small table) so readers can see all four strategies.
Because the four strategies are so close, the rest of this section uses max-pooling to isolate
the standalone contribution of Channel Boosting rather than of the fusion rule.

A companion ablation removes Channel Boosting entirely: BERT and RoBERTa encode the first
window independently and are combined by elementwise max-pooling, without the Channel Critic
Agent's trust weights or cross-gating (Table~\ref{tab:rawfusion}). This raw configuration
underperforms the BERT-only baseline by 0.40~pp accuracy and 0.19~pp F1, confirming that a
fixed combination rule, without per-document weighting of each encoder, can hurt performance
relative to a single well-tuned encoder (Section~\ref{sec:problem}). Adding Channel Boosting to
the same max-pool fusion recovers and exceeds the baseline, improving accuracy by 1.36~pp and
F1 by 1.46~pp over the raw configuration.

\begin{table}[!htbp]
\small
\centering
\caption{Channel Boosting's contribution, isolated (max-pool fusion)}\label{tab:rawfusion}
\begin{tabular}{@{}lcccc@{}}
\toprule
Configuration & Acc. & F1 & S-Recall & S-Prec. \\
\midrule
BERT (baseline) & 89.14\% & 89.33\% & 86.66\% & 92.21\% \\
Max-Pool, no Channel Boosting & 88.74\% & 89.14\% & 88.11\% & 90.19\% \\
\textbf{Max-Pool + Channel Boosting} & \textbf{90.10\%} & \textbf{90.60\%} & 91.20\% & 90.10\% \\
\bottomrule
\end{tabular}
\end{table}

\subsection{Effect of Class-Weighted Loss}\label{sec:classweight}

Table~\ref{tab:classweight} reports single-encoder and IC-MAS (Transformer fusion, fixed
rounds) results under unweighted and class-weighted ($w=[1.0,1.6]$) losses. Class weighting
alone raises BERT-only recall by 3.61~pp (86.66\%$\to$90.27\%) at a 3.23~pp precision cost, a
substantial, purely loss-driven effect. Unweighted IC-MAS nevertheless outperforms
class-weighted BERT-only on every metric (accuracy $+1.59$~pp, F1 $+1.51$~pp, recall
$+1.54$~pp, precision $+1.38$~pp), showing that the architecture's contribution is not an
artefact of cost-sensitive reweighting. Class weighting brings IC-MAS only a small further
recall gain (+0.32~pp) at a slight cost in accuracy and precision.

\begin{table}[!htbp]
\small
\centering
\caption{Effect of class-weighted loss}\label{tab:classweight}
\begin{tabular}{@{}lcccc@{}}
\toprule
Model & Acc. & F1 & S-Recall & S-Prec. \\
\midrule
BERT (unweighted) & 89.14\% & 89.33\% & 86.66\% & 92.21\% \\
BERT (weighted) & 88.98\% & 89.57\% & 90.27\% & 88.98\% \\
RoBERTa (unweighted) & 85.85\% & 86.51\% & 86.57\% & 86.45\% \\
RoBERTa (weighted) & 85.29\% & 86.44\% & 89.41\% & 83.76\% \\
IC-MAS (unweighted) & 90.57\% & 91.08\% & 91.81\% & 90.36\% \\
IC-MAS (weighted) & 90.48\% & 91.03\% & 92.13\% & 89.96\% \\
\bottomrule
\end{tabular}
\end{table}

\subsection{Adaptive Consultation Efficiency}\label{sec:adaptiveeff}

Table~\ref{tab:adaptive} crosses the fusion strategy with adaptive vs.\ fixed-round
consultation. Max-Pool fusion with adaptive consultation is the strongest configuration in
accuracy and F1, using 1.38 rounds on average, a 54.0\% reduction from the fixed 3-round
budget, with 93.0\% of documents halting early. Transformer fusion with adaptive consultation
halts slightly earlier (1.28 rounds) but at lower accuracy and precision. Because the encoder
passes are unchanged, these savings apply to the consultation stage only; the end-to-end
reduction in inference cost is correspondingly smaller.

\begin{table}[!htbp]
\small
\centering
\caption{Fusion strategy $\times$ consultation type}\label{tab:adaptive}
\begin{tabular}{@{}lccccc@{}}
\toprule
Configuration & Acc. & F1 & Recall & Prec. & Rounds \\
\midrule
Transformer, fixed & 90.57\% & 91.08\% & 91.81\% & 90.36\% & 3.00 \\
Transformer, adapt. & 90.40\% & 90.95\% & 92.07\% & 89.86\% & 1.28 \\
Max-Pool, fixed & 90.66\% & 91.11\% & 91.33\% & 90.90\% & 3.00 \\
\textbf{Max-Pool, adapt.} & \textbf{90.72\%} & \textbf{91.23\%} & 92.01\% & 90.46\% & 1.38 \\
\bottomrule
\end{tabular}
\end{table}

\subsection{Negative Results}\label{subsec:negative}

We report four negative results, since a design choice that failed is useful to future work
only if the likely reason is also understood.

(1)~\textit{Extending coverage beyond two windows.} Splitting each document into $N=4$ chunks
with generalized $N$-agent consultation (mean-aggregated messages) achieved only 89.26\%
accuracy and 89.75\% F1, lower than the two-window design on every metric, with a 2.06~pp
drop in recall. We attribute this to an over-smoothing effect analogous to that documented for
graph neural networks \cite{li2018}: as the number of interacting agents grows, each agent's
update summarizes messages from a larger neighbourhood, diluting the signal of any single
evidence-bearing region. With two agents, each update is dominated by one information-dense
counterpart; with four, it averages over three, and localized sensitivity evidence is washed
out before it reaches the decision. For Strategic~16K, the two-window split therefore appears
to capture the regions carrying most sensitivity-relevant evidence, and pairwise GRU-based
consultation outperforms its naive mean-message generalization.

(2)~\textit{Gradual MLP compression.} Replacing the sharply reducing Critic and Consensus MLPs
with a more gradual sequence of layers underperformed on all four metrics (accuracy
$-0.58$~pp) and showed substantially higher fold-to-fold variance. We interpret this as
over-parameterization relative to the available training signal: the sharp-reduction design
maps each input to a low-dimensional decision space in two or three layers, acting as a strong
regularizing prior, whereas the gradual variant adds parameters without adding label
information, increasing the capacity to memorize fold-specific noise. The narrower bottleneck
($1536 \to 256 \to 64 \to 2$ for the Consensus Agent) is therefore retained.

(3)~\textit{Attention-based gate generation.} Replacing the linear-plus-sigmoid gate in Gated
Channel Boosting with 8-head cross-attention (a 5$\times$ parameter increase in the boosting
component) reduced accuracy ($-0.18$~pp), F1 ($-0.19$~pp), and recall ($-0.40$~pp), with
negligible precision benefit. This suggests that the trust-and-gating decision is essentially
low-dimensional (how much one encoder's evidence should modulate the other's) and does not
require full multi-head interaction; the extra capacity again appears to hurt, consistent with
the over-parameterization pattern in (2).

(4)~\textit{First-window-doubled configuration.} Splitting the first 512 tokens into two
adjacent 256-token windows, in place of the first-and-last pair, underperformed the final
design on all four metrics, although it still clearly outperformed the BERT-only baseline
(accuracy $+1.18$~pp, Sensitive Recall $+4.80$~pp). This supports the evidence-location
pattern of Section~\ref{subsec:evidencelocation}: reading the beginning twice forfeits the
concluding assessments and classification rationale found near the end of cables, and extra
detail from the early portion does not compensate.
% TODO(author): consider reporting all four metrics for this ablation in a small table.

Together, these results point to a consistent lesson: on a corpus of this size, narrower and
more constrained design choices, and reading the beginning and end of a document rather than
the beginning twice, generalize better than more flexible alternatives.

\subsection{Decision \& Verification Agent: Results}\label{subsec:decisionresults}

Of the disagreements between the Classification and Domain Evidence Agents across the
16{,}000 test predictions, 289 (1.8\% of the collection) pass the document-length gate
(Section~\ref{sec:decision}) and are sent to the Decision Agent. Table~\ref{tab:decision}
reports the resulting performance alongside the BERT baseline and the core IC-MAS
Classification Agent.

\begin{table}[!htbp]
\small
\centering
\caption{Performance progression from the BERT baseline through the core IC-MAS to the full MAS}\label{tab:decision}
\begin{tabular}{@{}lcccc@{}}
\toprule
Configuration & Acc. & F1 & S-Recall & S-Prec. \\
\midrule
BERT baseline & 89.14\% & 89.33\% & 86.66\% & 92.21\% \\
Core IC-MAS (Classification Agent) & 90.72\% & 91.23\% & 92.01\% & 90.46\% \\
\textbf{Full MAS (+ Decision Agent)} & \textbf{91.32\%} & \textbf{91.87\%} & \textbf{93.54\%} & 90.27\% \\
\bottomrule
\end{tabular}
\end{table}

Relative to the core Classification Agent, accuracy increases by 0.60~pp, F1 by 0.64~pp, and,
most importantly given the priority on minimizing missed sensitive documents
(Section~\ref{subsec:protocolmetrics}), Sensitive Recall by 1.53~pp, at a precision cost of
only 0.19~pp. This is the only tested Decision Agent configuration that improves accuracy, F1,
and recall over the core architecture; alternative trigger and weighting schedules did not
(Section~\ref{sec:decisionablations}).

\begin{figure}[!htbp]
\centering
\includegraphics[width=0.85\textwidth]{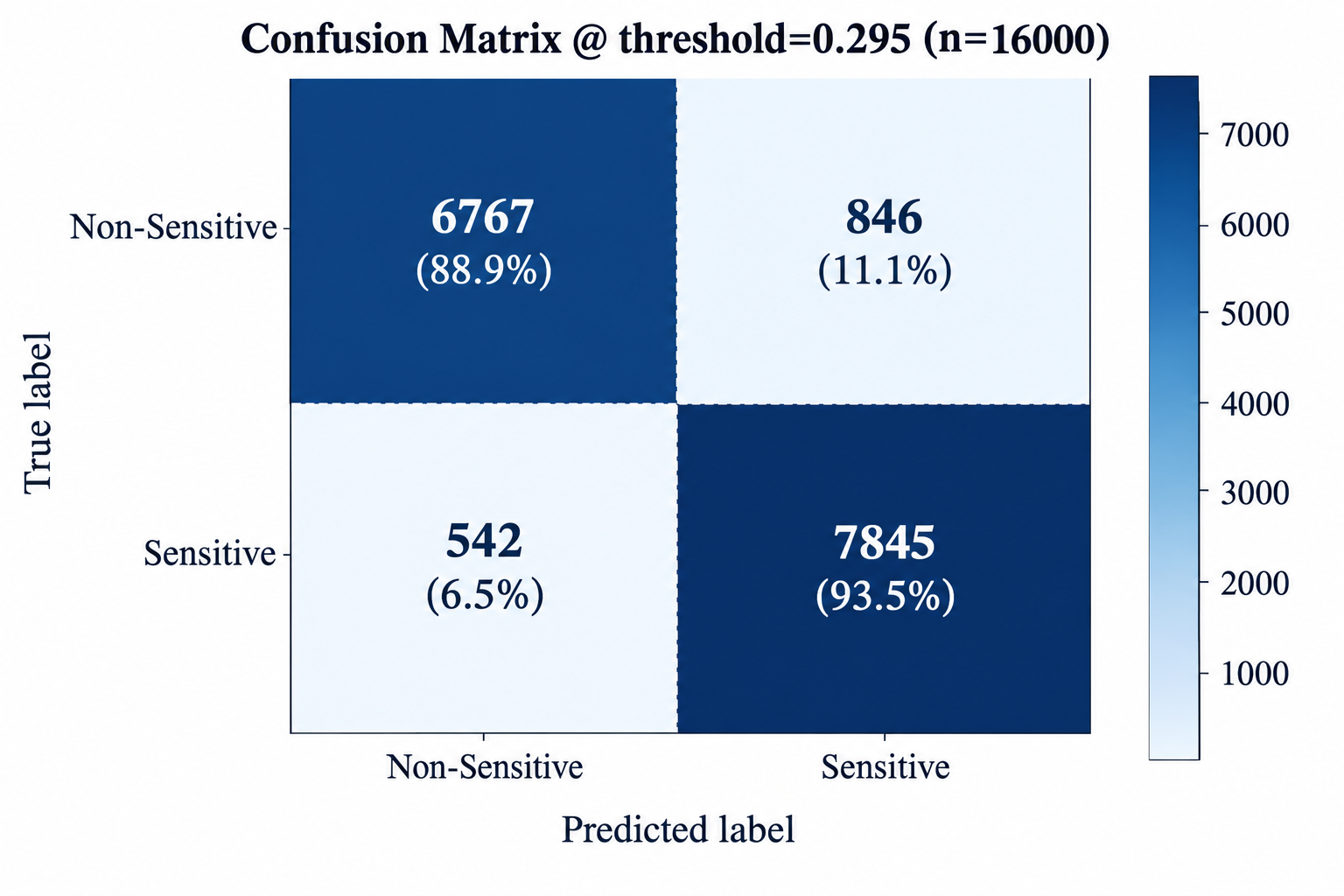}
\caption{Confusion matrix for the full MAS across all 16{,}000 documents.}\label{fig:confmatfull}
\end{figure}

\begin{figure}[!htbp]
\centering
\includegraphics[width=\textwidth]{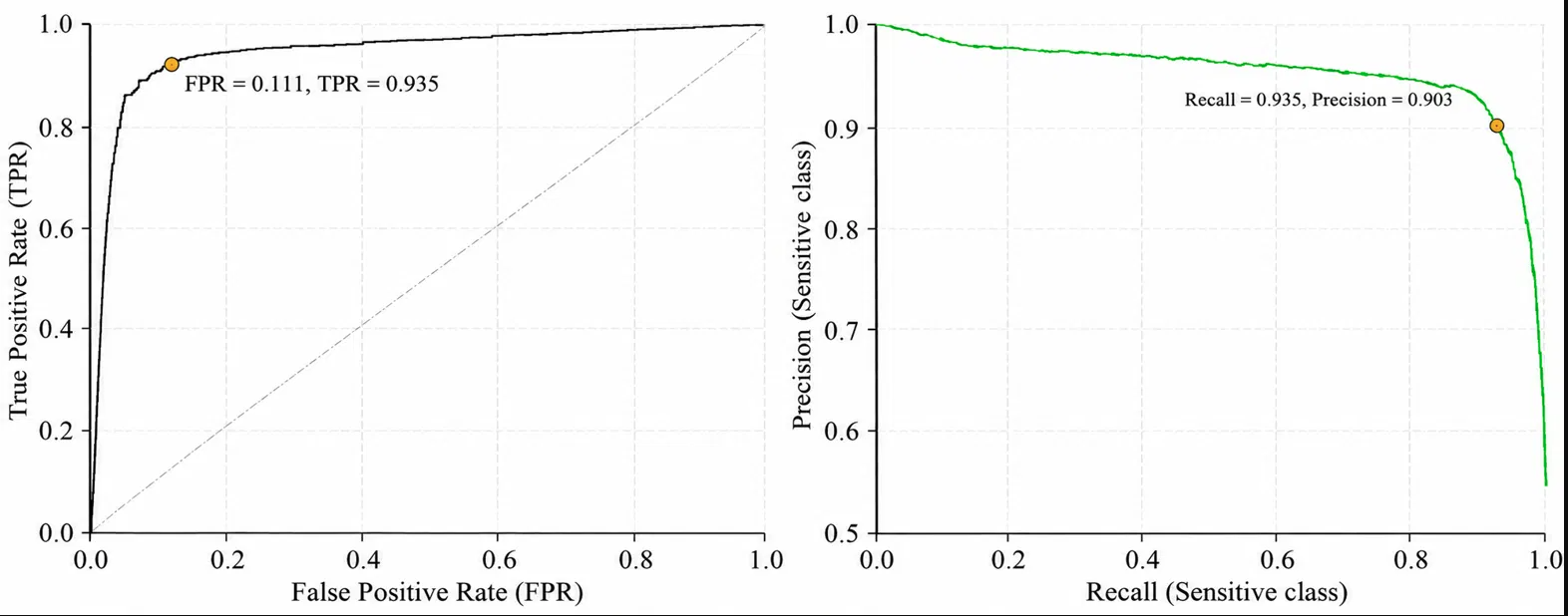}
\caption{Full MAS ROC (left) and precision--recall (right) curves, pooled across all 16{,}000 documents.}\label{fig:rocfull}
\end{figure}

Figures~\ref{fig:confmatfull} and~\ref{fig:rocfull} show the full MAS confusion matrix and
ranking curves, for comparison with the core model's results in Figures~\ref{fig:confmat}
and~\ref{fig:roc}.

To test whether this improvement could arise by chance on the small subset of documents the
Decision Agent acts on, McNemar's test was applied to the pooled per-document predictions of
the full MAS and the core Classification Agent ($n=16{,}000$; Table~\ref{tab:mcnemarfull}).

\begin{table}[!htbp]
\small
\centering
\caption{McNemar contingency table, full MAS vs.\ core IC-MAS}\label{tab:mcnemarfull}
\begin{tabular}{@{}lcc@{}}
\toprule
 & Core IC-MAS correct & Core IC-MAS wrong \\
\midrule
Full MAS correct & 14{,}515 & 90 \\
Full MAS wrong & 6 & 1{,}389 \\
\bottomrule
\end{tabular}
\end{table}

Of the 96 documents on which the two configurations differ, consistent with the small subset
the Decision Agent acts on, the full MAS is correct on 90 and the core Classification Agent on
only 6 ($\chi^2=71.76$, $p=2.43\times10^{-17}$). The per-document predictions used for this test
come from a separate inference run, so the pooled accuracies implied by
Table~\ref{tab:mcnemarfull} differ from the fold-averaged figures in Table~\ref{tab:decision}
by at most 0.04~pp, reflecting run-to-run variation. The Decision \& Verification Agent's
improvement is therefore systematic rather than a favourable point estimate.

\subsubsection{Ablations}\label{sec:decisionablations}

A purely statistical trigger rule (based on evidence strength alone, without the
document-length gate) never fired, since no disagreement met the required strength threshold.
Two few-shot in-context-learning variants of the reasoning step reached only 39.67\% and
11.50\% accuracy on held-out conflict subsets, the more aggressive variant being strongly
biased towards predicting \textsc{sensitive}. Removing the document-length gate reproduced the
same collapse, with 95\% of disagreements arising in documents with no unseen section. These
results indicate that grounding the Decision Agent's reasoning in retrieved, region-specific
text at decision time, rather than a fixed rule or a small set of examples, is key to its
benefit.
% TODO(author): give the core classifier's accuracy on the same conflict subsets as the
% reference point; the overall 89--91% accuracy is not comparable to accuracy on conflicts.

\section{Statistical Validation and Explainability}\label{sec:stats}

\subsection{Statistical Significance Testing}\label{sec:stattest}

The core configuration was compared with BERT-only using McNemar's test \cite{mcnemar1947} on
pooled per-document predictions ($n=16{,}000$) and a paired $t$-test on the five per-fold
scores. Of the 1{,}042 documents on which the two models disagreed, IC-MAS was correct on 643
and BERT-only on 399 ($\chi^2=56.67$, $p=5.15\times10^{-14}$), rejecting the null hypothesis
of equal error rates.
% TODO(author): 643 - 399 = 244 net documents = 1.53 pp, but the accuracy gap in Table 4 is
% 1.58 pp (about 253 documents). Check that both come from the same BERT run.
At the fold level, the proposed model is significantly better than BERT-only in accuracy
($t=8.00$, $p=0.0013$), F1 ($t=6.43$, $p=0.0030$), and Sensitive Recall ($t=3.03$,
$p=0.0390$); the precision difference is not significant ($t=1.26$, $p=0.2753$). Because
fold-level scores from $k$-fold cross-validation share training data, standard paired $t$-tests
can overstate significance \cite{dietterich1998,nadeau2003}; we therefore treat McNemar's test
as the primary evidence and the fold-level tests as corroborating evidence with limited power.

\subsection{Explainability}\label{sec:xai}

The Explainability Agent tells a human reviewer, in plain language, \emph{why} a document
received its final class, after the decision has been made; it reads only the final class and
the document text and never influences the decision. It works in two stages. First, an
evidence-highlighting stage combines three complementary signals: keyword-based lexical
evidence, LIME \cite{ribeiro2016}, and SHAP \cite{lundberg2017}. Gradient-based attribution was
considered but rejected, because the shared BERT encoder is invoked twice per forward pass (once
per window), which would conflate the two calls' attribution paths under standard
layer-attribution tooling. LIME and SHAP, as black-box methods that query only the pipeline's
input--output behaviour, avoid this ambiguity. Given the cost of hundreds of forward passes per
explained document, this stage is demonstrated on 10 example documents from the held-out test
fold.
% TODO(author): consider adding one worked example (a short highlighted excerpt plus the
% generated explanation) as a figure; at present the XAI component has no visible output.

Second, an explanation-generation stage passes this evidence, together with the document text
and the assigned class, to Phi-3-mini-instruct, a lightweight instruction-tuned language model,
which writes a natural-language justification of the assigned class. The language model only
organizes and verbalizes evidence already produced by the keyword, LIME, and SHAP signals; it
receives no gradients or internal activations from the classifier and cannot alter its
prediction, so the black-box property is preserved. The generated explanations have not been
quantitatively evaluated (e.g., for fidelity to the underlying model or for usefulness to human
reviewers); both the highlighted evidence and the language-model explanations should therefore
be read as diagnostic aids rather than a causal account of the model's decisions.

\section{Discussion and Limitations}\label{sec:discussion}

\subsection{Why Consultation Outperforms OR-Logic}\label{sec:whyconsult}

OR-combination is a fixed operator on the precision--recall trade-off of the underlying models:
it can move the combined system's operating point along that trade-off, exchanging precision for
recall, but cannot push the trade-off curve outward. Iterative Consultation, by contrast, is a
learned function optimized jointly with the rest of the network, so the model can learn how much
weight a signal from one window should carry given the context provided by the other, a
context-sensitive form of evidence weighting that a fixed logical rule cannot express. The
results in Table~\ref{tab:main} are consistent with this view: consultation retains most of
OR-logic's recall gain while regaining 2.9--4.0~pp of precision relative to OR-combination.

\subsection{IC-MAS as a Cooperative Multi-Agent System}\label{subsec:coopmas}

We assess the architecture against standard criteria for a cooperative multi-agent system:
common goal (CG), task decomposition (TD), coordination, communication, and joint decision (JD)
(Table~\ref{tab:mas}). Every agent satisfies CG and TD by construction: the four core agents are
optimized jointly under a single loss, and the Domain Evidence and Decision \& Verification
Agents, though trained or configured separately, share the goal of correct sensitivity
classification. The Consultation Agents fully satisfy coordination and communication, since
each round's belief update is conditioned on an explicit message from the other agent; the
Consensus Agent fully satisfies JD, since its output depends jointly on both agents' final
beliefs. The Decision \& Verification Agent is the only agent that satisfies all five criteria:
it coordinates the two upstream prediction sources, communicates with the Domain Evidence Agent
by requesting region-specific evidence on \textsc{Investigate}, and its output is the final
joint decision when it is invoked. Among the core agents the system is cooperative but not
hierarchical: no core agent can override, select, or deactivate another. The Decision \&
Verification Agent is the exception, since it can override the Classification Agent's
prediction, but only on the small subset of documents described in
Section~\ref{subsec:decisionresults}.

\begin{table}[!htbp]
\small
\centering
\caption{Per-agent assessment against cooperative MAS criteria. \checkmark: satisfied;
$\sim$: partially satisfied. CG: common goal; TD: task decomposition; JD: joint decision.}\label{tab:mas}
\begin{tabular}{@{}lccccc@{}}
\toprule
Agent & CG & TD & Coord. & Comm. & JD \\
\midrule
Channel Critic Agent & \checkmark & \checkmark & \checkmark & $\sim$ & \\
Dual-Window Reader Agent & \checkmark & \checkmark & & & \\
Consultation Agents (A \& B) & \checkmark & \checkmark & \checkmark & \checkmark & $\sim$ \\
Consensus Agent & \checkmark & \checkmark & & & \checkmark \\
Domain Evidence Agent & \checkmark & \checkmark & & $\sim$ & \\
Decision \& Verification Agent & \checkmark & \checkmark & \checkmark & \checkmark & \checkmark \\
\bottomrule
\end{tabular}
\end{table}

\subsection{Limitations}\label{sec:limitations}
IC-MAS was evaluated on one corpus of diplomatic cables, and its performance on other
organisational documents remains unknown. The first-and-last-window design is supported by
ablation results, but the assumed evidence-location pattern has not been quantified. Direct
comparisons with long-context transformers, hierarchical models, and a head-plus-tail BERT
baseline are still needed to separate the benefits of wider coverage from those of consultation.
Other limitations include document-level supervision, limited fold-level statistical power,
poorly calibrated confidence estimates, and explanations not yet validated by human reviewers.
Future evaluations should prioritize these comparisons, confidence calibration, human assessment
of explanations, and testing on diverse document types and finer-grained sensitivity labels.

\section{Conclusion}\label{sec:conclusion}
This paper introduced IC-MAS, a Channel-Boosted Multi-Agent System for document sensitivity
classification. It combines critic-guided BERT--RoBERTa fusion with iterative consultation
between agents reading the beginning and end of a document. On Strategic~16K, the core model
achieved 90.72\% accuracy, 91.23\% F1, and 92.01\% sensitive recall, outperforming the BERT
baseline while requiring 54\% fewer consultation rounds than fixed-round consultation.
Adding full-document evidence and selective verification increased sensitive recall to 93.54\%.
Ablations indicate that adaptive channel fusion accounts for much of the improvement, while
consultation helps reconcile evidence without the precision loss of simple OR-combination.

By bridging classic deep learning mechanics with the strategic orchestration of compound AI
systems, this work introduces Channel Boosted Intelligence. Our architecture demonstrates how
treating diverse agentic capabilities as parallel informational channels enables robust,
nuanced document sensitivity classification. As the artificial intelligence landscape
transitions from monolithic models toward collaborative multi-agent ecosystems, Channel
Boosted Intelligence offers both a unifying vocabulary and a principled framework to guide
this evolution with clarity and rigor. Evaluation on other document collections and against
long-context baselines remains necessary to establish how far this framework generalizes.

\section*{CRediT authorship contribution statement}

\textbf{Aleesha Zainab:} Conceptualization, Methodology, Software, Investigation, Formal analysis,
Writing -- original draft. \textbf{Asifullah Khan:} Conceptualization, Supervision, Methodology, Writing -- review \&
editing. \textbf{Muhammad Ahmed Khalid:} Supervision, Writing -- review \& editing.
\textbf{Faheem Ullah Khan:} Supervision, Writing -- review \& editing.
% TODO(author): adjust roles to match the CRediT taxonomy as you see fit.

\section*{Declaration of competing interest}

The authors declare that they have no known competing financial interests or personal
relationships that could have appeared to influence the work reported in this paper.

% Author action required: Elsevier requests disclosure of substantive generative-AI
% assistance in manuscript preparation. Confirm tool(s), purpose, and author review;
% replace the placeholders below with a truthful statement before submission.
\section*{Declaration of generative AI and AI-assisted technologies in the writing process}
During the preparation of this work, the authors used [name of tool/service] to
[describe the writing or editing assistance]. The authors reviewed and edited the
resulting text and take full responsibility for the content of the publication.

\section*{Funding}

This research did not receive any specific grant from funding agencies in the public,
commercial, or not-for-profit sectors.

\section*{Data availability}

The Strategic~16K corpus is derived from the publicly available WikiLeaks Public Library of US
Diplomacy (PlusD). Code is available from the corresponding author on reasonable request.
% TODO(author): consider releasing the list of cable identifiers and fold assignments so that
% the corpus can be reconstructed without redistributing the documents themselves.
% TODO(author): reviewers may raise the use of leaked classified material. Consider a short
% statement that only publicly released PlusD content was used, for research purposes.

% References follow Elsevier numbered citation order. Verify each source metadata/DOI before submission.


\begin{thebibliography}{00}

\bibitem{zainab2026benchmark} A. Zainab, M. A. Khalid, F. U. Khan, and A. Khan, Benchmarking Classical and Transformer-Based Models for Document Sensitivity Classification, arXiv preprint arXiv:2608.16928, 2026.
\bibitem{ibm2023} IBM Security, Cost of a Data Breach Report 2023, IBM Corporation, 2023.
\bibitem{alzhrani2019} K. Alzhrani, F. S. Alrasheedi, F. A. Kateb, and T. E. Boult, CNN with Paragraph to Multi-Sequence Learning for Sensitive Text Detection, in Proc. 2nd Int. Conf. Comput. Appl. Inf. Secur. (ICCAIS), 2019, pp. 1--6.
\bibitem{petrolini2022} M. Petrolini, S. Cagnoni, and M. Mordonini, Automatic Detection of Sensitive Data Using Transformer-Based Classifiers, Future Internet, vol. 14, no. 8, art. 228, 2022.
\bibitem{mcdonald2019} G. McDonald, A Framework for Technology-Assisted Sensitivity Review: Using Sensitivity Classification to Prioritise Documents for Review, Ph.D. dissertation, School of Computing Science, Univ. of Glasgow, 2019.
\bibitem{minaee2021} S. Minaee, N. Kalchbrenner, E. Cambria, N. Nikzad, M. Chenaghlu, and J. Gao, Deep Learning Based Text Classification: A Comprehensive Review, ACM Comput. Surv., vol. 54, no. 3, pp. 1--40, 2021.
\bibitem{devlin2019} J. Devlin, M.-W. Chang, K. Lee, and K. Toutanova, BERT: Pre-training of Deep Bidirectional Transformers for Language Understanding, in Proc. NAACL-HLT, 2019, pp. 4171--4186.
\bibitem{ahmad2025} F. B. Ahmad, A. A. Kiani, Y. Hafeez, H. Imran, M. Habib, A. Nawaz, M. R. R. Rana, and M. Azhar, Securing Cloud Data: An Approach for Cloud Computing Data Categorization Based on Machine Learning, Int. J. Innov. Sci. Technol., vol. 7, no. 1, pp. 235--258, 2025.
\bibitem{minaee2025} S. Minaee et al., Large Language Models: A Survey, arXiv:2402.06196, 2025.
\bibitem{huang2022} Y. Huang, T. Lv, L. Cui, Y. Lu, and F. Wei, LayoutLMv3: Pre-training for Document AI with Unified Text and Image Masking, in Proc. 30th ACM Int. Conf. Multimedia, 2022, pp. 4083--4091.
\bibitem{gaspar2024} D. Gaspar, P. Silva, and C. Silva, Explainable AI for Intrusion Detection Systems: LIME and SHAP Applicability on Multi-Layer Perceptron, IEEE Access, vol. 12, 2024.
\bibitem{rudin2019} C. Rudin, Stop Explaining Black Box Machine Learning Models for High Stakes Decisions and Use Interpretable Models Instead, Nature Mach. Intell., vol. 1, no. 5, pp. 206--215, 2019.
\bibitem{saritha2025} P. S. Saritha and R. Kumar, Sensitive Data Protection Using AI: An Evaluation of Deep Learning and Metaheuristic-Based Leakage Prevention Techniques, in Proc. Int. Conf. Intell. Commun. Netw. Comput. Techn. (ICICNCT), 2025, pp. 1--6.
\bibitem{hart2011} M. Hart, P. Manadhata, and R. Johnson, Text Classification for Data Loss Prevention, in Privacy Enhancing Technologies (PETS), LNCS vol. 6794, Springer, 2011, pp. 18--37.
\bibitem{jacobs1991} R. A. Jacobs, M. I. Jordan, S. J. Nowlan, and G. E. Hinton, Adaptive Mixtures of Local Experts, Neural Comput., vol. 3, no. 1, pp. 79--87, 1991.
\bibitem{li2016} Y. Li, D. Tarlow, M. Brockschmidt, and R. Zemel, Gated Graph Sequence Neural Networks, in Proc. ICLR, 2016.
\bibitem{li2018} Q. Li, Z. Han, and X.-M. Wu, Deeper Insights into Graph Convolutional Networks for Semi-Supervised Learning, in Proc. AAAI Conf. Artif. Intell., 2018.
\bibitem{graves2016} A. Graves, Adaptive Computation Time for Recurrent Neural Networks, arXiv:1603.08983, 2016.
\bibitem{liu2019} Y. Liu et al., RoBERTa: A Robustly Optimized BERT Pretraining Approach, arXiv:1907.11692, 2019.
\bibitem{cho2014} K. Cho et al., Learning Phrase Representations Using RNN Encoder--Decoder for Statistical Machine Translation, in Proc. EMNLP, 2014.
\bibitem{ribeiro2016} M. T. Ribeiro, S. Singh, and C. Guestrin, ``Why Should I Trust You?'' Explaining the Predictions of Any Classifier, in Proc. ACM SIGKDD, 2016.
\bibitem{lundberg2017} S. M. Lundberg and S.-I. Lee, A Unified Approach to Interpreting Model Predictions, in Proc. NeurIPS, 2017.
\bibitem{du2024} Y. Du, S. Li, A. Torralba, J. B. Tenenbaum, and I. Mordatch, Improving Factuality and Reasoning in Language Models Through Multiagent Debate, in Proc. ICML, 2024.
\bibitem{survey2025} R. Alva Principe et al., Long Document Classification in the Transformer Era: A Survey on Challenges, Advances, and Open Issues, WIREs Data Min. Knowl. Discov., 2025.
\bibitem{schuster2022} T. Schuster et al., Confident Adaptive Language Modeling, in Proc. NeurIPS, 2022.
\bibitem{fedus2022} W. Fedus, B. Zoph, and N. Shazeer, Switch Transformers: Scaling to Trillion Parameter Models with Simple and Efficient Sparsity, J. Mach. Learn. Res., vol. 23, no. 120, pp. 1--39, 2022.
\bibitem{hayesroth1985} B. Hayes-Roth, A Blackboard Architecture for Control, Artif. Intell., vol. 26, no. 3, pp. 251--321, 1985.
\bibitem{mcnemar1947} Q. McNemar, Note on the Sampling Error of the Difference Between Correlated Proportions or Percentages, Psychometrika, vol. 12, no. 2, pp. 153--157, 1947.
\bibitem{dietterich1998} T. G. Dietterich, Approximate Statistical Tests for Comparing Supervised Classification Learning Algorithms, Neural Comput., vol. 10, no. 7, pp. 1895--1923, 1998.
\bibitem{nadeau2003} C. Nadeau and Y. Bengio, Inference for the Generalization Error, Mach. Learn., vol. 52, no. 3, pp. 239--281, 2003.
\bibitem{clark2020} K. Clark, M.-T. Luong, Q. V. Le, and C. D. Manning, ELECTRA: Pre-training Text Encoders as Discriminators Rather Than Generators, in Proc. ICLR, 2020.
\bibitem{sanh2019} V. Sanh, L. Debut, J. Chaumond, and T. Wolf, DistilBERT, a Distilled Version of BERT: Smaller, Faster, Cheaper and Lighter, arXiv preprint arXiv:1910.01108, 2019.
\bibitem{zhou2023} X. Zhou, T. Zhang, C. Cheng, and S. Song, Dynamic Multichannel Fusion Mechanism Based on a Graph Attention Network and BERT for Aspect-Based Sentiment Classification, Appl. Intell., vol. 53, pp. 6800--6813, 2023.
\bibitem{liu2024} C. Liu, Y. Wang, and J. Yang, A Transformer-Encoder-Based Multimodal Multi-Attention Fusion Network for Sentiment Analysis, Appl. Intell., vol. 54, pp. 8415--8441, 2024.
\bibitem{kumar2024} A. Kumar and D. Toshinwal, HLC: Hierarchically-Aware Label Correlation for Hierarchical Text Classification, Appl. Intell., vol. 54, pp. 1602--1618, 2024.
\bibitem{tsirmpas2024} D. Tsirmpas, I. Gkionis, G. Th. Papadopoulos, and I. Mademlis, Neural Natural Language Processing for Long Texts: A Survey on Classification and Summarization, Eng. Appl. Artif. Intell., vol. 133, art. 108231, 2024, doi:10.1016/j.engappai.2024.108231.
\bibitem{zhao2025shapg} C. Zhao, J. Liu, and E. Parilina, ShapG: New Feature Importance Method Based on the Shapley Value, Eng. Appl. Artif. Intell., vol. 148, art. 110409, 2025, doi:10.1016/j.engappai.2025.110409.
\bibitem{kosztyan2025} Z. T. Koszty\'an, T. Kir\'aly, T. Csizmadia, A. I. Katona, and \'A. Vathy-Fogarassy, Automated Research Methodology Classification Using Machine Learning, Eng. Appl. Artif. Intell., vol. 156, art. 111039, 2025, doi:10.1016/j.engappai.2025.111039.
\bibitem{ferraro2025} A. Ferraro, G. M. Orlando, and D. Russo, Generative Agent-Based Modeling with Large Language Models for Insider Threat Detection, Eng. Appl. Artif. Intell., vol. 157, art. 111343, 2025, doi:10.1016/j.engappai.2025.111343.
\bibitem{alevizos2026} V. Alevizos, N. Gerolimos, D. Georgantzelis, and G. A. Papakostas, A Large Language Model-Based Multi-Agent Framework for Sustainable Industrial Design, Eng. Appl. Artif. Intell., vol. 176, art. 114818, 2026, doi:10.1016/j.engappai.2026.114818.
\bibitem{dantas2024} P. V. Dantas, W. S. da Silva Jr., L. C. Cordeiro, and C. B. Carvalho, A Comprehensive Review of Model Compression Techniques in Machine Learning, Appl. Intell., vol. 54, pp. 11804--11844, 2024.

\end{thebibliography}
\end{document}